\documentclass[10pt,twocolumn,letterpaper]{article}

\usepackage[pagenumbers]{iccv} % To force page numbers, e.g. for an arXiv version

\usepackage{fontawesome}
\usepackage{multirow}
\usepackage{arydshln}
\usepackage{comment}

\newlength\savewidth\newcommand\shline{\noalign{\global\savewidth\arrayrulewidth
		\global\arrayrulewidth .8pt}\hline\noalign{\global\arrayrulewidth\savewidth}}

\definecolor{iccvblue}{rgb}{0.21,0.49,0.74}
\usepackage[pagebackref,breaklinks,colorlinks,allcolors=iccvblue]{hyperref}

\def\paperID{2250} % *** Enter the Paper ID here
\def\confName{ICCV}
\def\confYear{2025}

\definecolor{lightgray}{gray}{0.9}
\newcommand{\CLA}[1]{{\color[HTML]{E76254} \textbf{#1}}}
\newcommand{\CLB}[1]{{\color[HTML]{4472c4} \textbf{#1}}}

\title{Information Density Principle for MLLM Benchmarks}

\author{Chunyi Li$^{1,}$$^{2}$$^{*}$, Xiaozhe Li$^{2,}$$^{3}$$^{*}$, Zicheng Zhang$^{1,}$$^{2}$, Yuan Tian$^{1,}$$^{2}$, Ziheng Jia$^{1}$,\\ Xiaohong Liu$^{1}$, Xiongkuo Min$^{1}$, Jia Wang$^{1}$, Haodong Duan$^{2}$$^{\dag}$, Kai Chen$^{2}$, Guangtao Zhai$^{1,}$$^{2}$$^{\dag}$\\
Shanghai Jiao Tong University$^{1}$, Shanghai AI Lab$^{2}$, Tongji University$^{3}$\\
}
\begin{document}

\twocolumn[{%
\renewcommand\twocolumn[1][]{#1}%
\maketitle
\begin{center}
    \centering
    \vspace{-2.4em}
    \includegraphics[width=1\linewidth]{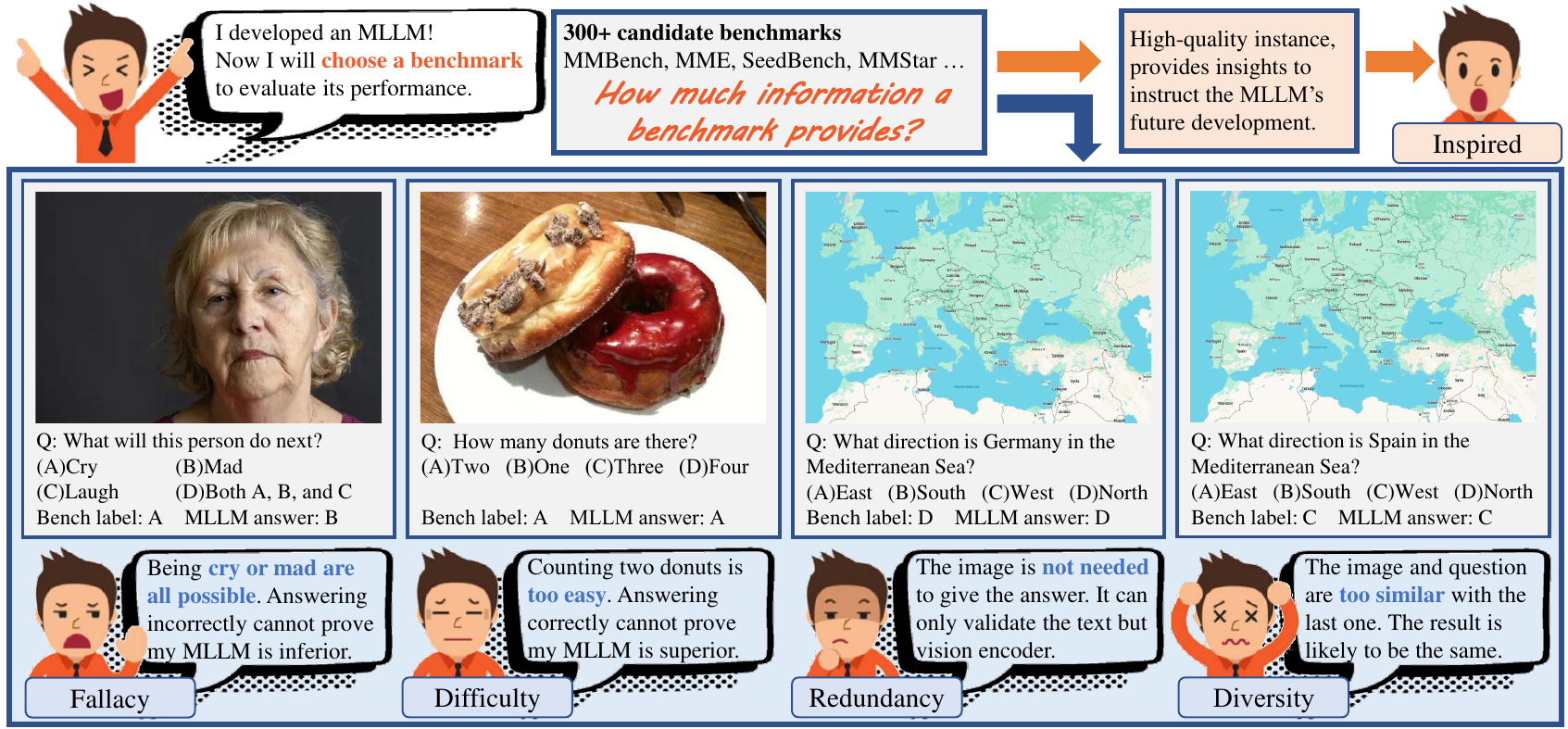}
    \vspace{-20pt}
    \captionof{figure}{The significance of the information density for the MLLM research community, which is a principle for selecting suitable benchmarks. Validating MLLMs on a \CLA{high information density} benchmark can provide insight for their future development; However, \CLB{low information density} benchmarks give less useful information, or the information is not reliable, due to the four defects above.} 
    \label{fig:spotlight}
    \vspace{-0.5mm}
\end{center}%
}]

\begin{abstract}

With the emergence of Multimodal Large Language Models (MLLMs), hundreds of benchmarks have been developed to ensure the reliability of MLLMs in downstream tasks. However, the evaluation mechanism itself may not be reliable. For developers of MLLMs, questions remain about which benchmark to use and whether the test results meet their requirements. Therefore, we propose a critical principle of Information Density, which examines how much insight a benchmark can provide for the development of MLLMs. We characterize it from four key dimensions: (1) Fallacy, (2) Difficulty, (3) Redundancy, (4) Diversity. Through a comprehensive analysis of more than 10,000 samples, we measured the information density of 19 MLLM benchmarks. Experiments show that using the latest benchmarks in testing can provide more insight compared to previous ones, but there is still room for improvement in their information density. We hope this principle can promote the development and application of future MLLM benchmarks. Project page: \href{https://github.com/lcysyzxdxc/bench4bench}{https://github.com/lcysyzxdxc/bench4bench}.
\end{abstract}

\vspace{-5mm}
\section{Introduction}
\label{sec:intro}

Evaluation plays a crucial role in driving the development of Multimodal Large Language Models (MLLM) as it reveals the strengths and weaknesses of models across various content and dimensions, thereby guiding their future optimization directions. Currently, the MLLM research community primarily uses benchmarks, which involve a series of questions with pre-annotated answers, to conduct in-depth and comprehensive evaluations of MLLM. According to statistics from Opencompass, there are currently 300+ MLLM benchmarks, with 100+ already integrated into various multimodal evaluation suites. \cite{platform:chatbot,platform:lmms-eval,platform:vlmevalkit,platform:EvalScope,platform:flagevalmm} Given the abundance of benchmarks, it is impossible for MLLM developers to test all. Thus, it poses a significant challenge to choose suitable benchmarks that meet their requirements.

For MLLM developers, a benchmark serves the MLLM, and a good benchmark should provide as many insights as possible, revealing where the model needs improvement. Therefore, we define the quality of this service as \CLA{Information Density: the volume of meaningful information reflected by a certain number of samples}. There are four attributes that lead to defects in information density:

\begin{itemize}
    \item Fallacy: A sample is poorly formulated, where the information reflected is not reliable.
    \item Difficulty: A sample is too simple thus almost all MLLMs are correct, giving no meaningful information.
    \item Redundancy: A sample that can be answered correctly based on only part of the information, with the remaining part being redundant.
    \item Diversity: Multiple samples that are too similar, resulting in overlapping information reflected in the responses.
\end{itemize}

As shown in Figure \ref{fig:spotlight}, the information density principle can help MLLM developers in filtering through numerous benchmarks to avoid the above four defects, thereby providing insights for the development of MLLM. Based on such principle, our contributions are concluded as follows:

\begin{itemize}
    \item We interpreted the information density from information entropy mathematically, thus establishing a theoretical basis for `What an insightful benchmark should be'.
    \item We evaluated 19 mainstream benchmarks with 17,912 instances, namely a `benchmark for benchmark' characterizing the capabilities of the above four dimensions, providing selection references for MLLM developers.
    \item We provided a Human-Model-Data evaluation pipeline, supporting end-to-end input of benchmark datasets, and output of information density in different dimensions, providing design references for benchmark developers.
\end{itemize}

\section{Related Works}

\subsection{MLLM Benchmarks}
MLLM benchmarks include two categories: one is Visual Question Answering (VQA) \cite{vqa:mmmu,vqa:mmvet}, which allows for open-ended textual responses; the other is Multiple Choice Question (MCQ) \cite{data:A-okvqa,data:MMBench}, which provides options within a pre-defined space. Generally, VQA is more challenging, but it has to compare the answer with human annotations during evaluation, making it difficult to quantify the similarity between these two sentences, leading to certain controversy \cite{eval:bleu,eval:cider,eval:spice} in the evaluation results. In contrast, MCQ maps continuous scores to two discrete outcomes (correct/incorrect), resulting in relatively objective and reliable outcomes. Therefore, considering reproducibility, mainstream evaluation leaderboards primarily use MCQ.

However, as shown in Figure \ref{fig:spotlight}, existing MCQ benchmarks have certain limitations, mainly in the following two aspects. First, due to the rapid development of MLLM, most benchmarks before 2023 \cite{data:MME,data:POPE} have been saturated and the difficulty is too low; for this reason, later benchmarks \cite{data:MME-RealWorld,data:Realworldqa} become more challenging, but such difficulty sometimes comes from unclear definitions of questions/options. Besides the correct option, other options are also reasonable. Meanwhile, there are two smaller limitations. One is the redundancy of content, namely the accidental leakage of answers. MLLMs may give the correct answer directly through text without using the image. Since MMstar \cite{data:MMstar} discovered this problem, current benchmarks have avoided it to some extent; the other is data overlap, namely multiple samples are asking the same. As a single instance, benchmarks may avoid three defects perfectly, but as a whole, they cannot reflect different attributes of MLLMs, resulting in the testing of many unnecessary \cite{bench:redundancy} samples.
These factors greatly restrict the development of MLLMs. Considering that benchmarks, as an evaluation mechanism, have never been evaluated, the above four attributes of benchmarks need to be revealed for the MLLM community.

\begin{table}[t]
\centering
    % \caption{Absolute robustness comparison between \textit{GPT-4o} and \textit{human} (\textbf{left/right}). Evaluated by 3 tasks, 3 strength, 7 steps, and 7 groups. As the R-bench champion, \textit{GPT-4o} still lags behind \textit{human} across the board.  \CLB{Orange}/\CLA{Blue} denote \textit{GPT-4o} performance below 90\% or above 98\% of \textit{humans}.}
    \caption{Three evaluation paradigms for Benchmarks. A paradigm has better accuracy, supports more defects, leading to more evaluation costs. {{\faCheckCircle}} denotes the supported defects during evaluation and {\CLA{\faCheckCircle}} is the default result for information density.}
    \label{tab:paradigm}
    \vspace{-8pt}
    \renewcommand\arraystretch{1.4}
    \belowrulesep=0pt\aboverulesep=0pt
%     \resizebox{\linewidth}{!}{
% \begin{tabular}{lcccc}
% Evaluator  & Fallacy                           & Difficulty                        & Redundency                        & Diversity                         \\
% Human Eval & {\CLA{\faCheckCircle}} & {\faCheckCircle}                                 & {\faCheckCircle}                                 & {\faCheckCircle}                                 \\
% Model Eval &                                   & {\CLA{\faCheckCircle}} & {\CLA{\faCheckCircle}} & {\CLA{\faCheckCircle}} \\
% Data Eval  &                                   & {\faCheckCircle}                                 &                                   & {\faCheckCircle}                                
% \end{tabular}}
    \resizebox{\linewidth}{!}{
\begin{tabular}{l|c|cccc}
\toprule
Evaluator  & Cost & Fallacy                           & Difficulty                        & Redundancy                        & Diversity                         \\ \midrule
Human Eval & $\uparrow\uparrow\uparrow$ & {\CLA{\faCheckCircle}} & {\faCheckCircle}                                 & {\faCheckCircle}                                 & {\faCheckCircle}                                 \\
Model Eval & $\uparrow\uparrow$ &                                   & {\CLA{\faCheckCircle}} & {\CLA{\faCheckCircle}} & {\CLA{\faCheckCircle}} \\
Data Eval  & $\uparrow$ &                                   & {\faCheckCircle}                                 &                                   & {\faCheckCircle} \\ \bottomrule                               
\end{tabular}}
\end{table}

\subsection{Evaluation for Benchmarks}
\label{sec:b4b}

Some early-stage works have discovered various problems with the benchmark. LIME \cite{bench:lime} emphasis on the trade-off between Fallacy and Difficulty. For samples that MLLM gets wrong, manual verification is used to determine whether the sample is indeed challenging or not rigorous. However, this paradigm is `Human Eval', which requires high costs during annotation. Zerobench \cite{bench:zero} and Enigmaeval \cite{bench:enigmaeval} focus on Difficulty and Redundancy, without involving human subjects. However, they all adopt `Model Eval', using the performance of MLLMs to reflect the data quality. For example, the most advanced model scores less than 10\% \cite{model:deepseek,model:internvl2,model:internvl25,model:qwen2,model:qwenvl25,model:mplugowl3} can prove the benchmark is difficult enough. There is still a lack of a `Data Eval' pipeline without using MLLMs for inference, which directly judges the benchmark defects based on its content.

\section{Evaluation Framework}
\subsection{Supported Paradigms}

As summarized in section \ref{sec:b4b}, we have three evaluation paradigms as shown in Table \ref{tab:paradigm}. Among these, Human Eval scores the dataset through a human expert panel, Model Eval scores based on the inference of MLLM on the dataset, and Data Eval assigns a score directly to the data. Regarding the four possible defects of the benchmark, Fallacy requires determining whether the sample is correctly formulated, which inevitably involves human annotation; Redundancy refers to whether the question can be answered correctly in the absence of certain information, which must be accomplished by either humans or models. Additionally, the Difficulty and Diversity dimensions support a complete Human-Model-Data paradigm. Since humans are the golden standard for evaluation, the accuracy of the three decreases in turn, but the cost can be saved successively. Thus, to provide an automated evaluation pipeline, information density is first selected for Model Eval, then Data Eval, and finally Human Eval. The performance of Model/Data Eval is measured by their correlation with Human Eval.

\subsection{Theoretical Basis}
\label{sec:theory}

According to the definition in section \ref{sec:intro}, the amount of information \cite{add:entropy-1,add:entropy-2,add:entropy-3} in a single instance is represented as $I$, and the information density of the entire dataset is its expectation ${\rm E}(I)$, which is specifically expressed as follows:

\begin{equation}
  \begin{aligned}
    {\rm E}(I) &= {\rm E}(J) \cdot \frac{I \cdot J}{||I|| \cdot ||J||} \cdot D_{div}\\
               &= {\rm E}(J) \cdot (1-D_{red}) \cdot D_{div} ,
  \end{aligned}
\end{equation}
where ${\rm E}(J)$ is the absolute information expectation of $J$. Since the information from $J$ is not necessarily insightful to the subjects, it will be weakened on two levels. At the instance level, for a subject MLLM, absolute information should overlap with the information it is concerned about, while the orthogonal part is redundant and recorded as $D_{red}$; at the dataset level, different samples of the same dataset may be very similar. For a benchmark, it may be possible to measure completely consistent indicators using only some samples. This diversity limitation is noted as $D_{div}$. Then, the information ${\rm E}(I_0)$ can be computed as:

%% 冗余度red要不要改，div如何加sigma符号

\begin{equation}
  \begin{aligned}
    {\rm E}(J) &= - m {\rm log_2}({m_0}) \cdot {\rm C}(m)-n {\rm log_2}(n_0) \cdot {\rm C}(n)\\
               &= \epsilon + n \cdot {\rm log_2}(\frac{1}{n_0}) \cdot {\rm C}(n)\\
               &\approx D_{dif} \cdot {\rm log_2}(\frac{1}{n_0}) \cdot (1-D_{fal}),
  \end{aligned}
\end{equation}
where $m,n$ represents the proportions of correct/incorrect samples for MLLM inference, while $m_0,n_0$ denotes the expected proportions of correct/incorrect responses in the research community according to the information entropy model. Obviously, $m + n =m_0+n_0 = 1$. Before 2023, getting a question right was a low-probability event, which is why achieving $m$ generated more information. However, as MLLM gradually becomes an expert today, answering correctly has turned into a high-probability event, resulting in $n_0$ becoming much smaller than $m_0$. Consequently, only when MLLM feels challenged, indicating the occurrence of $n$, can more information be obtained. The confidence function $\rm C (\cdot)$ indicates the credibility of the question, expressed as the proportion of rigorous samples among all samples. Since $m$ and $n$ are variables, given (1) the information generated by $n_0$ is inherently greater than $m_0$, and the $log$ function further amplifies this disparity, and (2) the confidence ${\rm C} (m,n)$ from the same dataset cannot be several times different, the term about $m$ will be an ignorable infinitesimal $\epsilon$ compare to the term about $n$. Here, $n$ itself represents difficulty $D_{dif}$, while ${\rm C} (n)$ is the proportion of remaining samples after excluding fallacy samples $D_{fal}$, Thus:

\begin{equation}
   {\rm E}(I) \propto (1-D_{fal}) \cdot D_{dif} \cdot (1-D_{red}) \cdot D_{div},
\end{equation}
where we decompose the abstract concept of `insight' into a multiplicative combination of four dimensions. This theory will serve as the basis for the quantitative assessment of defects and the conclusions drawn in subsequent experiments.

\begin{table}[t]
\centering
    % \caption{Absolute robustness comparison between \textit{GPT-4o} and \textit{human} (\textbf{left/right}). Evaluated by 3 tasks, 3 strength, 7 steps, and 7 groups. As the R-bench champion, \textit{GPT-4o} still lags behind \textit{human} across the board.  \CLB{Orange}/\CLA{Blue} denote \textit{GPT-4o} performance below 90\% or above 98\% of \textit{humans}.}
    \caption{An example of difficult/easy cases for \textbf{Difficulty} evaluation in the data level, according to the Structure (Image), Grammar (Text), Option (Text), and Region (Image+Text).}
    \label{tab:difficulty}
    \vspace{-8pt}
    \renewcommand\arraystretch{1.4}
    \belowrulesep=0pt\aboverulesep=0pt
    \resizebox{\linewidth}{!}{
\begin{tabular}{lp{4.7cm}p{4.7cm}}
\toprule
Dimension    & \multicolumn{1}{c}{Difficult Case}                                                                                        & \multicolumn{1}{c}{Easy Case}                                                                    \\ \midrule
Structure     & Image with detailed elements, rich textures, and colors. & Image with clear composition, limited textures, and colors. \\
Grammar     & Question: Which transportation on the bustling street suggests the presence of modern technology? & Question: What is the style of the image?                               \\
Option     & Options: (A) Self-driving cars, (B) Electric vehicles, (C) Drones have similar semantics.                        & Options: (A) Busy street, (B) Ocean waves, (C) Mountain view have different meanings. \\
Region & `Transportation' refers to a specific small region of the image.                        & `Style' refers to the overall appearance and theme of the image.        \\ \bottomrule
\end{tabular}}
\vspace{-3mm}
\end{table}

\section{Evaluation Pipeline}

This section will introduce the evaluation method of each defect, following the Human-Model-Data order listed in Table \ref{tab:paradigm}. Among them, Fallacy only supports Human Eval, which is the default Infomation Density result. For the other three, to achieve automated evaluation, we only use Human Eval as the ground truth label, while Model Eval serves as the Infomation Density result. The goal of Model/Data Eval is to match the results of Human Eval.

\subsection{Difficulty}

As summarized in section \ref{sec:theory}, for a sample in the dataset, Difficulty is the prior conditions determining Fallacy. Thus we introduce this paradigm first. 
For Human Eval, we organize a group of human experts to rate the difficulty from 0 to 5 (with a granularity of 0.5), using 100 random samples of a benchmark. The total score of a benchmark was obtained by averaging all experts/all samples. This result is not reported as Infomation Density, but to verify whether the Model/Data Eval is reasonable.

\begin{figure}[tb]
\centering
\includegraphics[width = \linewidth]{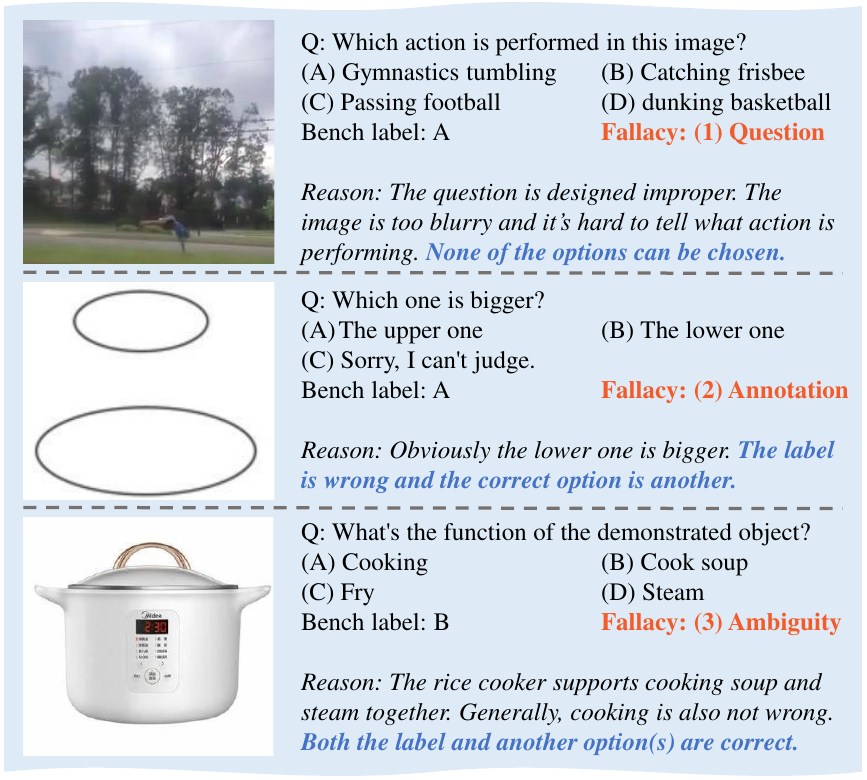}
\vspace{-7mm}
\caption{\textbf{Fallacy} cases for the benchmark original label. For the original label and other options, excepting the label is correct and others are incorrect, the other three correctness situations make up the cases. [Keys: \CLA{Sub-type}, \CLB{Definition}]}
\vspace{-4mm}
\label{fig:fallacy}
\end{figure}

For Model Eval, we reflect this indicator through the correctness of responses by the MLLM. We apply the most commonly used proprietary MLLM, GPT-4o \cite{model:gpt4}, as well as two leading open-source models on Opencompass, InternVL-2.5 \cite{model:internvl25}, and QwenVL-2.5 \cite{model:qwenvl25}, requiring the models to predict the best option and a possible alternative. Based on the Mix-of-Expert (MoE) results of the three, we assess whether the difficulty of a sample is sufficient, categorized into the following three situations: 
\begin{itemize}
    \item Junior: The mean opinion from three models about the best option is incorrect.
    \item Extreme: All three models give incorrect best options.
    \item Ambiguity: The best and alternative options overlap in all models. (e.g. GPT-4o: B,C; QwenVL/InternVL: C,B)
\end{itemize}
where Extreme is the subset of Junior. After analyzing each sample (see supp-Table 2), we found that Ambiguity usually can be answered correctly, and only a few questions belong to both Junior and Ambiguity. Therefore, we believe they have no overlap in the dimension $D_{dif}$:

\begin{equation}
  D_{dif} = P(Q_{jun}) + P(Q_{amb}),
\end{equation}
where $Q_{jun},Q_{ext},Q_{amb}$ is the Junior, Extreme and Ambiguity sub-dimensions while $P(\cdot)$ denotes their probability.

For Data Eval, samples can be directly analyzed to determine their difficulty without the inference of MLLMs. Table \ref{tab:difficulty} presents examples of an easy and a difficult sample, from which the four characteristics of difficult samples can be easily identified as: 
\begin{itemize} 
    \item Structure: For images, the structural information should be rich to examine the analytical ability of MLLMs. 
    \item Grammar: For text, the grammar should be complex to assess the comprehension ability of MLLMs. 
    \item Option: For text, the semantics of the options should be as close as possible to test the confusion of MLLMs. 
    \item Region: For combinations of images and text, the subject of the question should focus on details within the image to evaluate the localization ability of MLLMs. 
\end{itemize} 
which can be computed as follows: (1) textural information from the 2D Laplacian operator, (2) depth of the syntax \cite{tool:nltk} tree, (3) averaged CLIP distance \cite{tool:clip} between options, and (4) entropy of the syntax root node \cite{tool:clip-sur,tool:sam} in the image. Finally, through learnable parameters, we perform a linear combination of these factors to closely fit the $D_{dif}$ result of Model Eval, ensuring effective Data Eval.

\subsection{Fallacy}

After a sample is considered difficult enough, to ensure its credibility, we organized experts to mark all suspicious samples in $D_{dif}$. (See supp-Table 3). In the three cases in Figure \ref{fig:fallacy}, the sample will be defined as Fallacy. If the original label is correct, there must be other correct ones (Ambiguity) that lead to Fallacy; otherwise, there are either other potential choices (Option), or none of them are correct (Question). Therefore, $D_{fal}$ can be obtained:

\begin{equation}
  D_{fal} = P((Q_{que}+Q_{ano}+Q_{amb})|D_{dif} = 1 ),
\end{equation}
where $Q_{que},Q_{ano},Q_{amb}$ is the sub-dimensions above with no intersection, which can be directly added under the prior that the question has been answered incorrectly by MLLM.

\begin{figure}[tb]
\centering
\includegraphics[width = \linewidth]{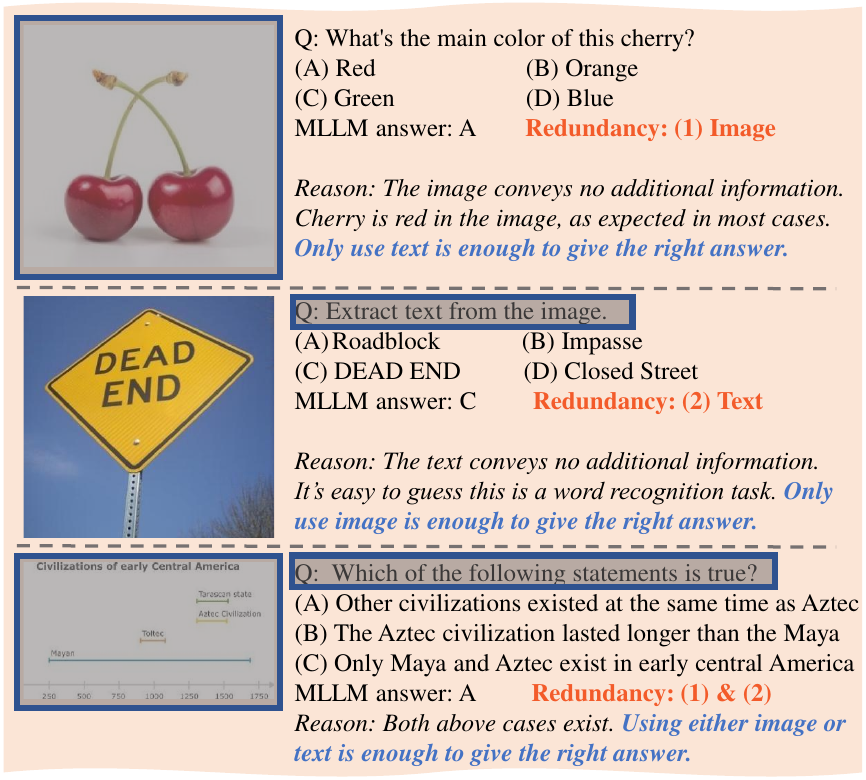}
\vspace{-7mm}
\caption{\textbf{Redundancy} cases for the MLLM answer. If MLLM can give correct inference when image/text information is invisible, then this discarded part is redundant. [Keys: \CLA{Sub-type}, \CLB{Definition}, \colorbox{gray!20}{Discarded infomation}]}

\vspace{-4mm}
\label{fig:redundancy}
\end{figure}

\subsection{Redundancy}

In contrast to Difficulty, Redundancy considers the situation of getting the right answer even when some information is discarded as Figure \ref{fig:redundancy} shows. If the sample is common sense, or the question itself reveals the answer, the correct inference can be given using only the text, and the image conveys no information; if the sample question is worthless, then the MCQ will degenerate into a caption task, and the correct inference can be given using only the image, and the text has no insight. 
For Human Eval, we organize similar subjective annotations for the two defects above, allowing humans to judge `whether the information is sufficient to give the correct answer' in the absence of image/text. Based on the human perception mechanism of different modalities, these two sub-dimensions are labeled respectively rather than together, by averaging all samples.

\begin{figure}[tb]
\centering
\includegraphics[width = \linewidth]{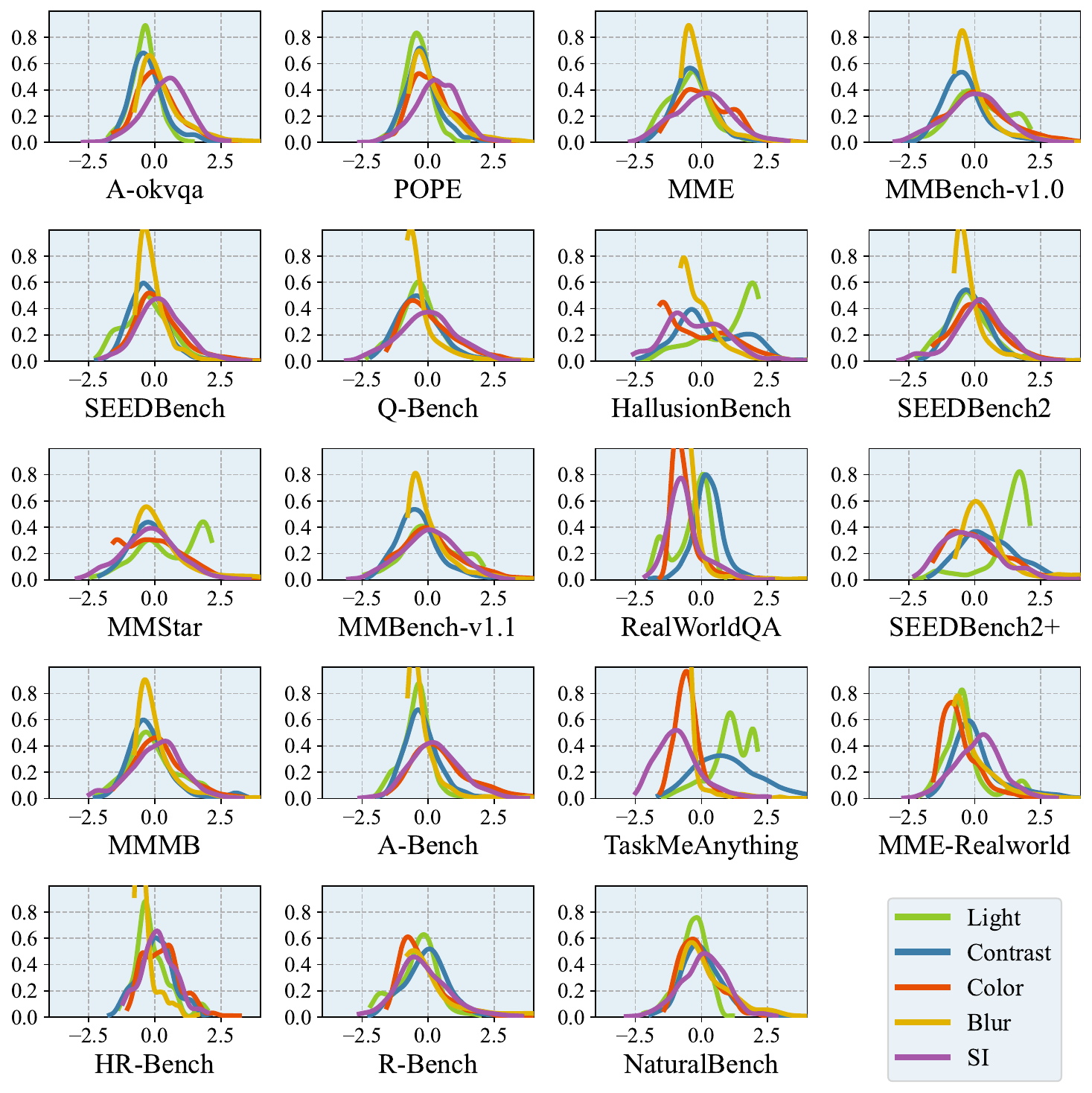}
\caption{Image \textbf{Diversity} in Data Eval, consists of five low-level attributes. Their distribution is merged to match the Model Eval. Benchmarks with a wide distribution have strong diversity, while benchmarks with a flat distribution tend to have similar content.}
\vspace{-4mm}
\label{fig:diversity-img}
\end{figure}

For Model Eval, We proceed with inference in the absence of image/text. Noted that although Redundancy and Fallacy are all negative cases, we do not negate the entire sample, but only exclude the redundant part in $D_{red}$:

\begin{equation}
  D_{red} =  \frac{w_{img}\cdot {\rm Acc}(\overline{I_{img}}) + w_{txt}\cdot{\rm Acc}(\overline{I_{txt}})}{w_{img}+w_{txt}},
\end{equation}
where $(w_{img},w_{txt})$ is the weight of the image/text, which depends on the specific data structure of the benchmark (See supp-Table 4). Function $\rm Acc(\cdot)$ measures the inference accuracy when image or text information $(I_{img}, I_{txt})$ are deprived of samples.
Due to insufficient information, most of the MLLMs except QwenVL-2.5 \cite{model:qwenvl25} always refuse to answer, so we only use its inference result.

\begin{figure}[tb]
\centering
\includegraphics[width = \linewidth]{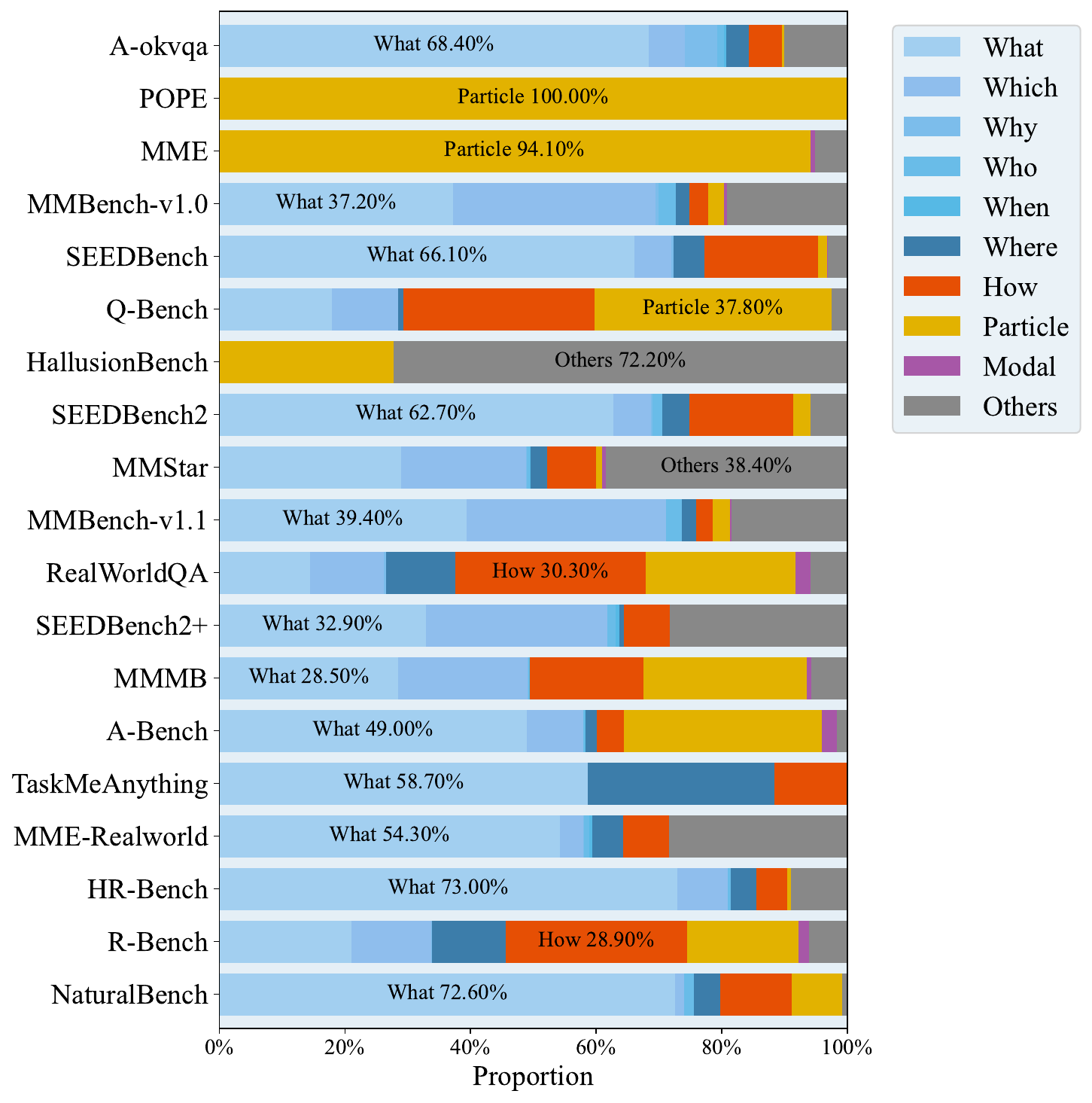}
\caption{Text \textbf{Diversity} in Data Eval, consists of ten first-word question formats. Their ratio is merged to match the Model Eval. Benchmarks with multiple question formats have greater diversity, while benchmarks asking the same format need improvement.}
\vspace{-4mm}
\label{fig:diversity-txt}
\end{figure}

\subsection{Diversity}

Diversity, just like Difficulty, can be analyzed from all three perspectives. Among these, Human Eval only serves as a validator; Model Eval allows a large-scale model to analyze samples sequentially, filtering out similar content, and reflecting diversity through the amount of remaining sample; Data Eval records the feature of each sample through a light-weighted algorithm, representing diversity through the breadth of their distribution. For Human Eval, Diversity is an index for the whole dataset rather than a single sample. Thus, after human experts complete the three tasks above (labeling one by one), they will give two overall diversity scores from 0 to 5 (with a granularity of 0.5) for all image/text samples, which was then averaged among experts.

\begin{figure*}[tb]
\centering
\begin{minipage}[]{0.24\linewidth}
  \centering
  \centerline{\includegraphics[width = \textwidth]{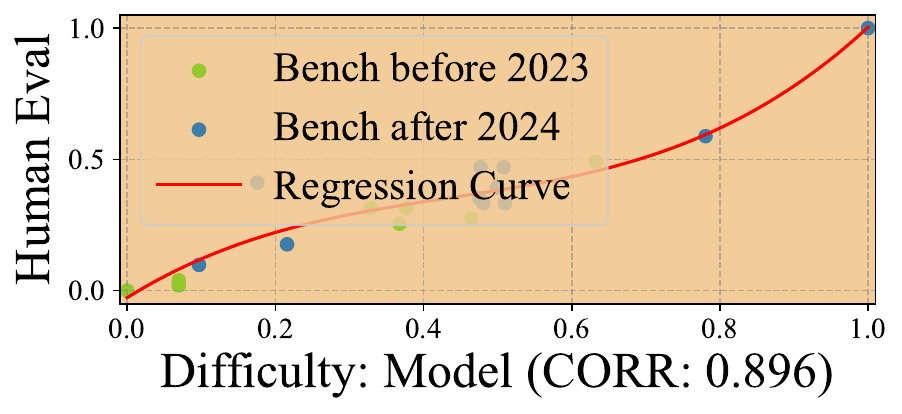}}
%  \centerline{(a)}\medskip
\end{minipage}
\begin{minipage}[]{0.24\linewidth}
  \centering
  \centerline{\includegraphics[width = \textwidth]{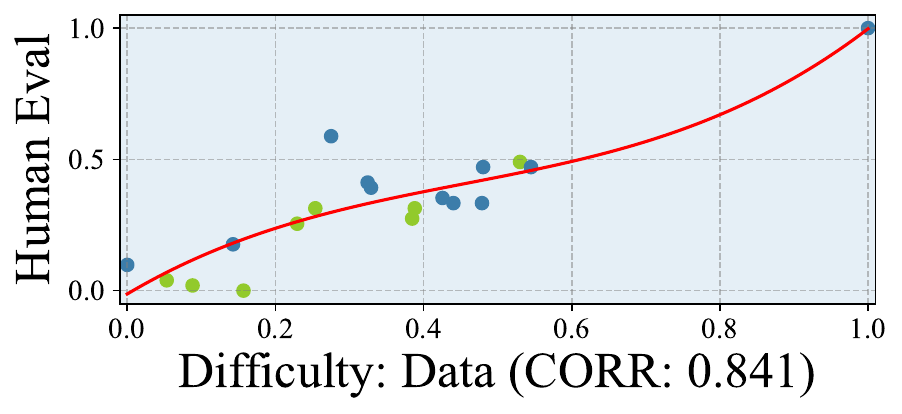}}
%  \centerline{(a)}\medskip
\end{minipage}
\begin{minipage}[]{0.24\linewidth}
  \centering
  \centerline{\includegraphics[width = \textwidth]{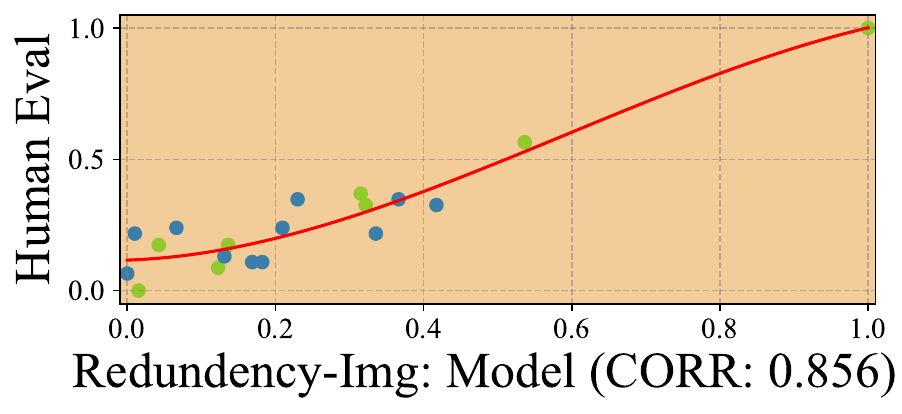}}
%  \centerline{(a)}\medskip
\end{minipage}
\begin{minipage}[]{0.24\linewidth}
  \centering
  \centerline{\includegraphics[width = \textwidth]{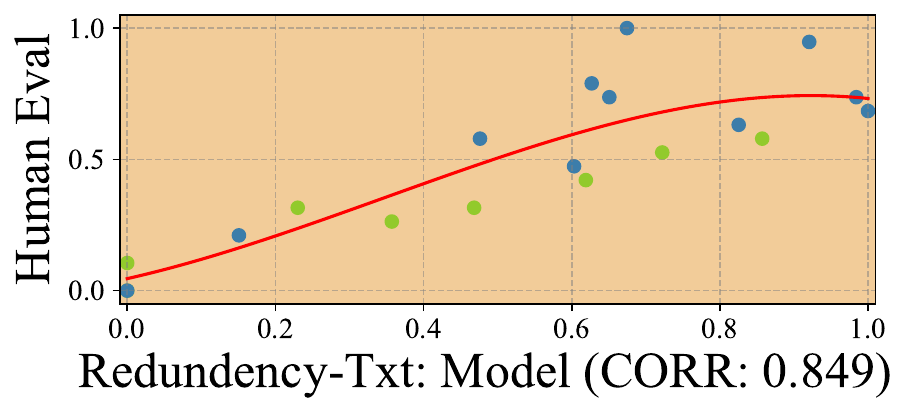}}
%  \centerline{(a)}\medskip
\end{minipage}
\begin{minipage}[]{0.24\linewidth}
  \centering
  \centerline{\includegraphics[width = \textwidth]{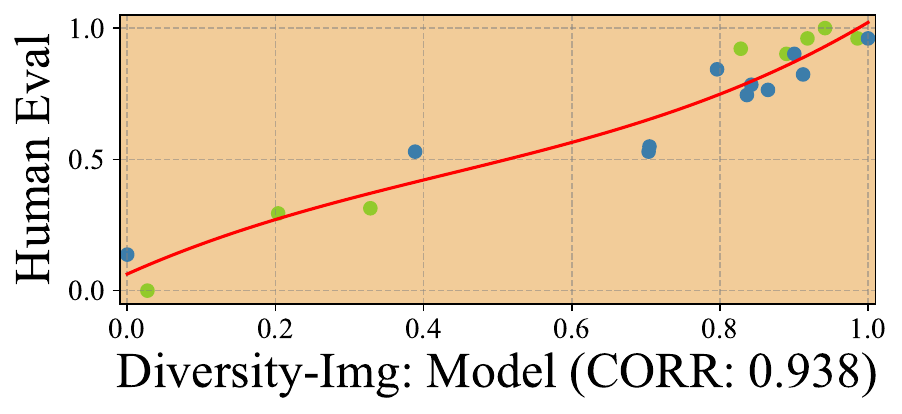}}
%  \centerline{(a)}\medskip
\end{minipage}
\begin{minipage}[]{0.24\linewidth}
  \centering
  \centerline{\includegraphics[width = \textwidth]{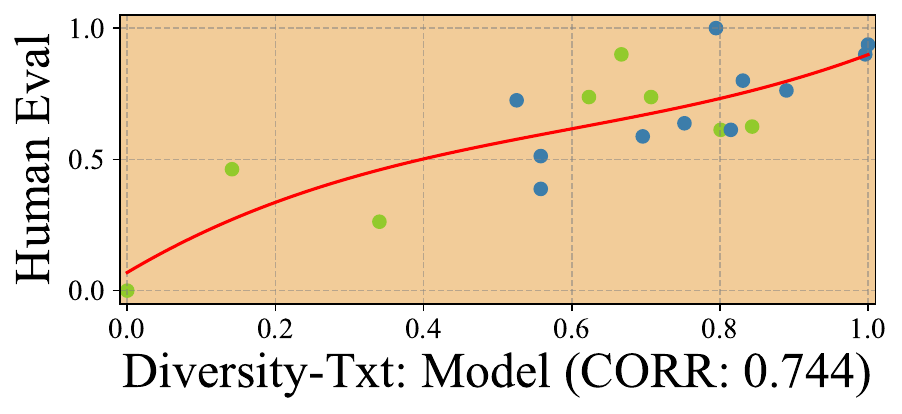}}
%  \centerline{(a)}\medskip
\end{minipage}
\begin{minipage}[]{0.24\linewidth}
  \centering
  \centerline{\includegraphics[width = \textwidth]{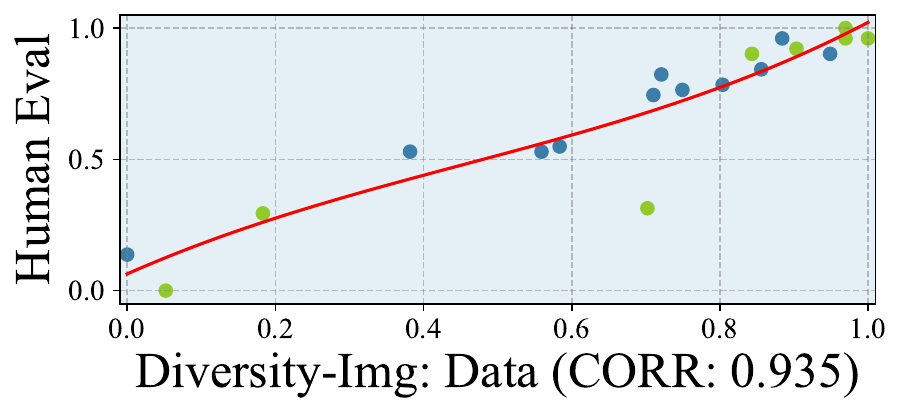}}
%  \centerline{(a)}\medskip
\end{minipage}
\begin{minipage}[]{0.24\linewidth}
  \centering
  \centerline{\includegraphics[width = \textwidth]{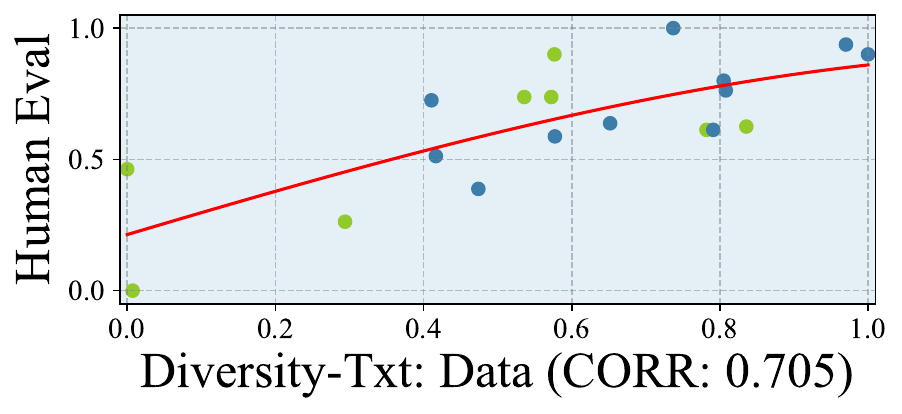}}
%  \centerline{(a)}\medskip
\end{minipage}
\vspace{-1mm}

\caption{The correlation between Model/Data Eval and Human Eval. These two evaluation results are consistent with human preferences with correlation coefficients above 0.7, which proves the rationality of the information density principle. [Keys: \CLA{Model Eval}; \CLB{Data Eval}]}
\vspace{-4mm}
\label{fig:human}
\end{figure*}

For Model Eval, we analyze each image and text sample using CLIP-image/text \cite{tool:clip} encoder, merging samples with high similarity, thus obtaining $D_{div}$:

\vspace{-2mm}
\begin{equation}
  D_{div} = \frac{w_{img}\cdot \frac{{\rm \#} ({\rm SIM}(I_{img}))}{{\rm \#} (I_{img})} + w_{txt}\cdot \frac{{\rm \#} ({\rm SIM}(I_{txt}))}{{\rm \#} (I_{txt})} } {w_{img}+w_{txt}},
\end{equation}
where ${\rm SIM}(\cdot)$ stands for the filtering function using k-means clustering, intra-cluster sorting, and semantic de-duplication \cite{method:semdedup} (see supp-Sec 3.4 for detailed implementation) above, and ${\rm \#}(\cdot)$ denotes the number of samples in a dataset. The more samples remain after filtering, the more diversity a dataset has. An image/text weight equalization is applied to combine two sub-dimensions into a final score.

For Data Eval, image and text are also conducted separately. For images, we analyze their low-level features, including: `Light' and `Contrast' as the average and difference of luminance. `Color' indicates the chrominance channel strength. `Blur' means the information density filtered by the Sobel operator. `SI' stands for the texture abundance. Previous works \cite{my:agiqa,my:aigiqa20k,my:misc} about human preference have demonstrated their correspondence to diversity, with specific calculation methods. The normalized distribution result is shown in Figure \ref{fig:diversity-img}, which is highly correlated to diversity. For example, Q-Bench \cite{data:Q-Bench} and A-Bench \cite{data:A-Bench} focus on low-level corruption, and the sharpness of the `Blur' curve indicates the coexistence of high/low-quality images, leading to a satisfying diversity. For texts, we categorize all questions into 10 types, including six special questions starting with `What/Which/Why/Who/When/Where', comprehensive questions starting with `How', particle questions starting with `Do/Does/Is/Are', modal verbs question started with `Can/Could/Should', and others.
In general, most benchmarks tend to ask `What' questions. Only three early benchmarks \cite{data:POPE,data:MME,data:Q-Bench} focus on Yes-or-No, most of which are about `Particles'; and two real-world benchmarks \cite{data:Realworldqa,data:R-bench} require executing various tasks, so they ask more `How' questions. Among them, the row for RealWorldQA \cite{data:Realworldqa} has various colors, implying highly diversified texts. Therefore, for a diverse benchmark, these 10 question types need to be covered widely and evenly.
Based on the above two conclusions in Data Eval, we use the variances of 5 image features to fit the image sub-dimension score, and the ratio of 10 text features to regress the text sub-dimension score in Model Eval. Thus, ensuring their combination $D_{div}$ of a benchmark can be represented by its image and text data components.

\section{Experiment}

\subsection{Candidate Benchmark}

Benchmarks that can be measured by the Information Density principle should meet three conditions: multiple-choice, multi-modal, and multi-domain. The specific applicability is described in the supp-Section 5. According to these conditions, we tested 19 mainstream benchmarks from 2022 to 2024, which are widely used in the research community to measure the performance of MLLM. These include (1) General topic: A-okvqa \cite{data:A-okvqa}, POPE \cite{data:POPE}, MME \cite{data:MME}, MMBench-v1.0 \cite{data:MMBench}, SEEDBench \cite{data:SEEDBench}, SEEDBench2 \cite{data:SEEDBench2}, MMStar \cite{data:MMstar}, MMBench-v1.1 \cite{data:MMBench}, SEEDBench2+ \cite{data:SEEDBench2+}, MMMB \cite{data:MMMB}, TaskMeAnything \cite{data:taskmeanything}, and HR-Bench \cite{data:HR-bench}; (2) Fixed image-topic: RealWorldQA (Real-world image)\cite{data:Realworldqa}, A-Bench (AI-Generated image) \cite{data:A-Bench}, MME-Realworld (Real-world image)\cite{data:MME-RealWorld}, and R-Bench (Distorted image)\cite{data:R-bench}; (3) Fixed text-topic: Q-Bench (Low-level question) \cite{data:Q-Bench}, HallusionBench (Hallusion question) \cite{data:Hallusionbench}, and NaturalBench (Naturalness question) \cite{data:naturalbench}. Except for benchmarks with only two options are not applicable for text Redundancy, all benchmarks are analyzed by the four defects about information density.

\begin{table*}[t]
\centering
    % \caption{Absolute robustness comparison between \textit{GPT-4o} and \textit{human} (\textbf{left/right}). Evaluated by 3 tasks, 3 strength, 7 steps, and 7 groups. As the R-bench champion, \textit{GPT-4o} still lags behind \textit{human} across the board.  \CLB{Orange}/\CLA{Blue} denote \textit{GPT-4o} performance below 90\% or above 98\% of \textit{humans}.}
    \caption{The information density for MLLM benchmarks in different dimensions, ordered by dataset released time. (The earliest version on Github, Huggingface, Arxiv, etc.) Abbreviations of Fallacy are for Question, Annotation, and Ambiguity; Difficulty are for Junior, Extreme, and Ambiguity. [Keys: \CLA{Best}; \CLB{Second Best}. Priority to the earlier benchmark under the same value.]}
    \label{tab:main}
    \vspace{-6pt}
    \renewcommand\arraystretch{1.4}
    \renewcommand\tabcolsep{7.5pt}
    \belowrulesep=0pt\aboverulesep=0pt
    \resizebox{\linewidth}{!}{
    % \tablestyle{8pt}{1.5}
% \begin{tabular}{|l|ll|ll|}
% \hline
% \multirow{2}{*}{} & \multicolumn{2}{l|}{}    & \multicolumn{2}{l|}{}    \\ \cline{2-5} 
%                   & \multicolumn{1}{l|}{} &  & \multicolumn{1}{l|}{} &  \\ \hline
%                   & \multicolumn{1}{l|}{} &  & \multicolumn{1}{l|}{} &  \\ \hline
%                   & \multicolumn{1}{l|}{} &  & \multicolumn{1}{l|}{} &  \\ \hline
% \end{tabular}
\begin{tabular}{l|c:ccc|c:ccc|c:cc|c:cc|r}
\shline
\multirow{2}{*}{Bench} & \multicolumn{4}{|c}{Fallacy $\downarrow$} & \multicolumn{4}{|c}{Difficulty $\uparrow$} & \multicolumn{3}{|c}{Redundency $\downarrow$} & \multicolumn{3}{|c}{Diversity $\uparrow$} & \multicolumn{1}{|c}{\multirow{2}{*}{Time}} \\ \cdashline{2-15} 
                       & \textbf{ALL} & Que   & Ano   & Amb   & \textbf{ALL}    & Jun   & Ext   & Amb   & \textbf{ALL}      & Img      & Txt      & \textbf{ALL}      & Img      & Txt     &                       \\ \shline
        A-okvqa \cite{data:A-okvqa} & 0.597 & 0.009 & 0.246 & 0.342 & 0.157 & 0.080 & 0.041 & 0.077 & 0.243  & 0.262  & 0.104   & 0.882 & 0.930 & 0.533 & Jun-2022 \\ 
        POPE \cite{data:POPE} & 0.557 & \CLA{0.000} & 0.557 & \CLA{0.000} & 0.119 & 0.099 & 0.041 & 0.020 & 0.562  & 0.634  & -   & 0.383 & 0.432 & 0.001 & May-2023 \\ 
        MME \cite{data:MME} & 0.526 & \CLB{0.000} & 0.466 & 0.060 & 0.206 & 0.133 & 0.060 & 0.073 & 0.133  & 0.153  & -   & 0.842 & 0.950 & 0.113 & Jun-2023 \\ 
        MMBench-v1.0 \cite{data:MMBench} & 0.578 & 0.028 & 0.174 & 0.376 & 0.157 & 0.093 & 0.047 & 0.064 & 0.149  & 0.154  & 0.126   & 0.861 & 0.906 & 0.674 & Jul-2023 \\ 
        SEEDBench \cite{data:SEEDBench} & 0.333 & 0.073 & 0.192 & 0.068 & 0.320 & 0.196 & 0.107 & 0.124 & 0.155  & 0.170  & 0.079   & 0.796 & 0.854 & 0.498 & Jul-2023 \\ 
        Q-Bench \cite{data:Q-Bench} & 0.280 & 0.000 & 0.268 & \CLB{0.012} & 0.373 & 0.203 & 0.098 & 0.170 & 0.175  & 0.182  & 0.116   & \CLA{0.951} & \CLB{0.987} & 0.640 & Sep-2023 \\ 
        HallusionBench \cite{data:Hallusionbench} & 0.269 & 0.000 & 0.103 & 0.166 & 0.465 & 0.306 & 0.138 & 0.159 & 0.312  & 0.362  & -   & 0.191 & 0.178 & 0.272 & Oct-2023 \\ 
        SEEDBench2 \cite{data:SEEDBench2} & 0.392 & 0.091 & 0.177 & 0.124 & 0.325 & 0.219 & 0.116 & 0.106 & 0.136  & 0.146  & 0.085   & 0.365 & 0.327 & 0.565 & Nov-2023 \\ 
        MMStar \cite{data:MMstar} & \CLA{0.135} & 0.021 & \CLA{0.079} & 0.035 & \CLB{0.546} & \CLB{0.362} & 0.182 & \CLA{0.184} & \CLB{0.054}  & \CLB{0.056}  & 0.045   & 0.827 & 0.866 & 0.664 & Mar-2024 \\ 
        MMBench-v1.1 \cite{data:MMBench} & 0.306 & 0.009 & 0.099 & 0.198 & 0.172 & 0.100 & 0.046 & 0.072 & 0.076  & 0.072  & 0.091   & 0.865 & 0.915 & 0.651 & Apr-2024 \\ 
        RealWorldQA \cite{data:Realworldqa} & 0.247 & 0.053 & 0.165 & 0.029 & 0.379 & 0.248 & 0.097 & 0.131 & 0.113  & 0.127  & \CLB{0.029}   & 0.756 & 0.749 & \CLB{0.796} & Apr-2024 \\ 
        SEEDBench2+ \cite{data:SEEDBench2+} & 0.646 & 0.128 & 0.319 & 0.199 & 0.397 & 0.291 & \CLB{0.200} & 0.106 & 0.252  & 0.292  & 0.082   & 0.818 & 0.861 & 0.635 & Apr-2024 \\ 
        MMMB \cite{data:MMMB} & 0.239 & 0.048 & 0.137 & 0.054 & 0.237 & 0.149 & 0.082 & 0.088 & 0.216  & 0.236  & 0.078   & 0.812 & 0.827 & 0.711 & Jun-2024 \\ 
        A-Bench \cite{data:A-Bench} & 0.333 & 0.023 & 0.205 & 0.105 & 0.398 & 0.223 & 0.101 & \CLB{0.175} & 0.214  & 0.232  & 0.108   & \CLB{0.941} & \CLA{0.999} & 0.601 & Jun-2024 \\ 
        TaskMeAnything \cite{data:taskmeanything} & \CLB{0.206} & 0.034 & \CLB{0.086} & 0.086 & 0.392 & 0.264 & 0.118 & 0.128 & 0.085  & 0.086  & 0.076   & 0.850 & 0.925 & 0.420 & Jun-2024 \\ 
        MME-Realworld \cite{data:MME-RealWorld} & 0.480 & 0.080 & 0.353 & 0.047 & \CLA{0.666} & \CLA{0.516} & \CLA{0.340} & 0.150 & \CLA{0.040}  & \CLA{0.047}  & \CLA{0.019}   & 0.701 & 0.750 & 0.556 & Aug-2024 \\ 
        HR-Bench \cite{data:HR-bench} & 0.369 & 0.079 & 0.166 & 0.124 & 0.380 & 0.302 & 0.142 & 0.078 & 0.113  & 0.124  & 0.060   & 0.205 & 0.155 & 0.446 & Aug-2024 \\ 
        R-Bench \cite{data:R-bench} & 0.336 & 0.106 & 0.176 & 0.054 & 0.382 & 0.255 & 0.111 & 0.127 & 0.110  & 0.119  & 0.059   & 0.873 & 0.885 & \CLA{0.799} & Sep-2024 \\ 
        NaturalBench \cite{data:naturalbench} & 0.464 & 0.073 & 0.179 & 0.212 & 0.215 & 0.117 & 0.054 & 0.098 & 0.209  & 0.244  & -   & 0.478 & 0.483 & 0.446 & Oct-2024 \\  \shline
\end{tabular}

}
\vspace{-4mm}
\end{table*}

\subsection{Experiment Setups}

We have adopted three mechanisms to ensure a fair evaluation: (1) Sampling alignment: For each benchmark, 1,000 samples are randomly selected. When the dataset is small, it is easy to ensure the high difficulty and credibility of individual questions, and there are almost no duplicate samples. However, large datasets inevitably include some low-quality samples, leading to insufficient information density. Aligning the data scale at the same level during evaluation helps eliminate interference and reflects the `real' data quality. (2) Circular evaluation: Using five different random seeds for MLLM evaluation, and says correct only if all five results match the original label. This effectively prevents random guessing, thereby reducing the false negative ratio of Difficulty and the false predicted ratio of Redundancy to negligible levels. (3) Anti-leakage: During the Data Eval and Model Eval processes, the results of Human Eval are not visible. The parameters of Data Eval can only refer to the results of Model Eval during training, while Model Eval directly outputs scores. This avoids overfitting humans to these two automated paradigms. Specifically, the features obtained by each benchmark in Data Eval are fitted to Model Eval using a random forest kernel with a maximum depth of 3, with the four features of Difficulty used together, while the five image features and ten text features of Diversity are trained separately to fit sub-dimension scores. 

For resource expenditure, Human Eval employs five Ph. D. candidates as an expert panel, whose extensive knowledge ensures the quality of the annotations; Model Eval utilizes 8$\times$ NVIDIA A800 SXM4 80GB GPUs to load the 72B parameter MLLM; and Data Eval only uses an Intel Xeon Platinum 8336C CPU for basic image/text processing.

\subsection{Human Trustworth Validation}

For the two evaluation paradigms, we calculate the correlation of the results from the two groups of evaluations, represented by the average of Spearman Rank-order Correlation Coefficient (SRCC) and Pearson Linear Correlation Coefficient (PLCC). A higher value represents a better correlation, and values above 0.7 indicate a sufficiently strong correlation\footnote{Equivalent to the same human subjects annotating twice.} between the two groups, while above 0.8 indicates a fundamental consistency.
Figure \ref{fig:human} shows the correlation between Model and Human Eval\footnote{For visualization, both axes have been normalized to 0-1, which does not affect SRCC and PLCC.}, including Difficulty, Redundancy sub-dimension, and Diversity sub-dimension. Overall, both automated evaluation methods sufficiently align with human subjective preference, with correlation coefficients above 0.7, proving the reliability during evaluation. For different dimensions, Difficulty, and Diversity-Img evaluations align most closely with human (about 0.9), followed by Redundancy-Img/Txt (about 0.85), and there is room for further improvement in Diversity-Txt evaluations. For different paradigms, Model Eval results are more accurate than Data Eval, but considering the lower cost of Data Eval, benchmark developers may choose based on their computational resources.

\subsection{Information Density Result and Discussion}

Table \ref{tab:main} presents the information density of each MLLM benchmark across various dimensions and sub-dimensions. The increase in Difficulty is a prevailing trend for benchmarks. In the context where MLLMs can solve various fundamental problems, it is necessary to design more specialized and challenging questions, specifically benchmarks with difficulty ratings above 0.3, or even 0.5, to ensure that the evolution of evaluations keeps pace with model advancements. However, excessively difficult questions often lead to Fallacy; under the premise that `MLLM finds this question is difficult', the Fallacy of benchmarks is nearly all above 0.2, with some exceeding 0.5, meaning that more than half of the questions answered incorrectly by MLLMs cannot yield any insights, whether due to the difficulty in discerning image details or the ambiguity of the question. Furthermore, the latest benchmarks are gradually avoiding the Redundancy flaw, while Diversity has not yet been overcome. This is particularly true for benchmarks aimed at specific tasks; the image components often involve molecular formulas (chemistry), evolutionary trees (biology), or maps (geography). MLLMs can answer these questions from specialized knowledge when information is incomplete. Meanwhile, the specialization also results in insufficient diversity in image and text samples. In summary, all four dimensions need improvement in future benchmarks.

\begin{figure}[tb]
\centering
\includegraphics[width = \linewidth]{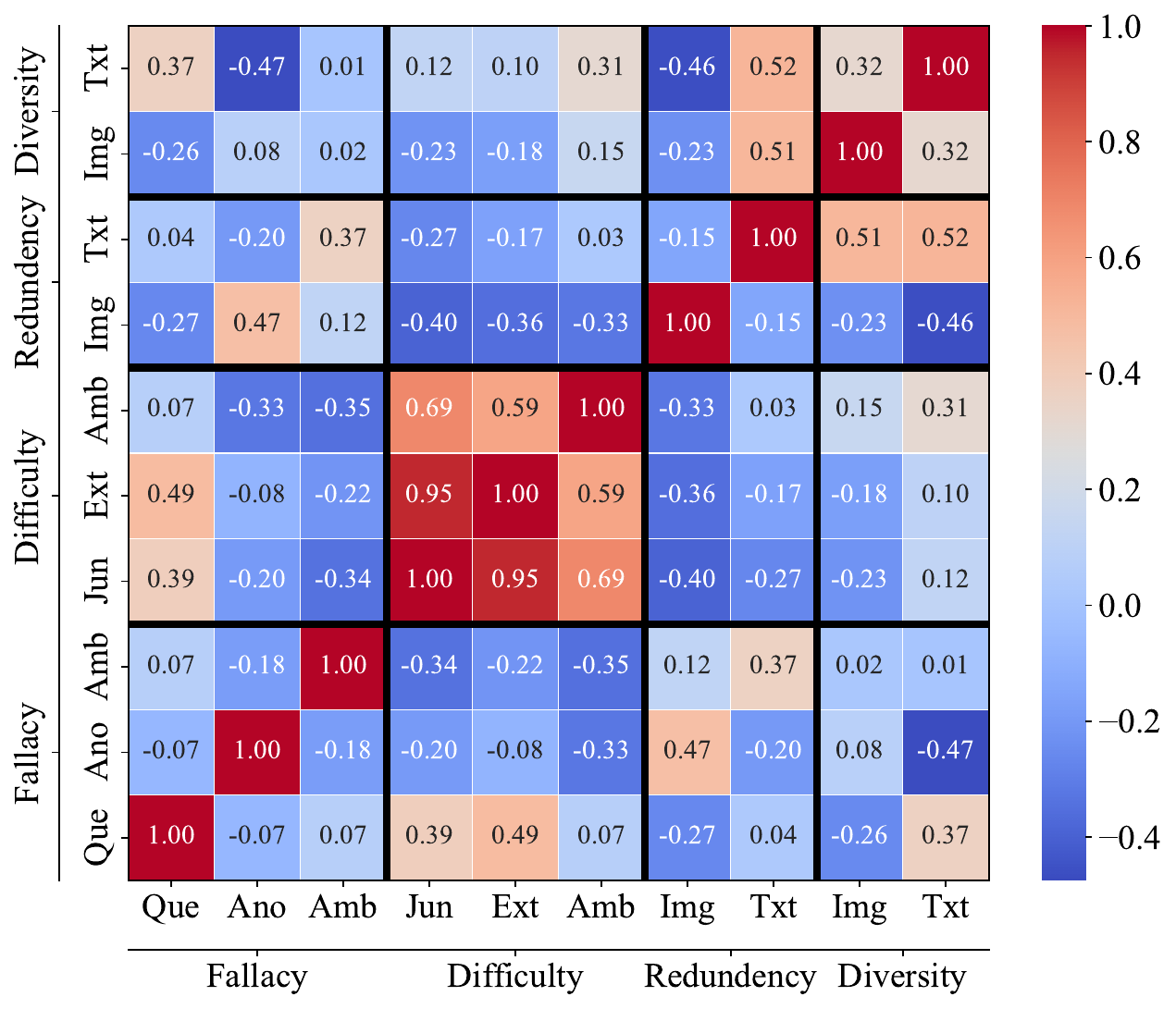}
\vspace{-7mm}
\caption{Correlation for MLLM benchmarks between different dimensions/sub-dimensions. Abbreviations follow Tabel \ref{tab:main}. Their low correlation proves the rationality of the division.}
\vspace{-6mm}
\label{fig:corrmat}
\end{figure}

An interesting finding is the `mutual antagonism' between dimensions. For early-stage benchmarks, all dimensions can be improved together at a low baseline. However, for the latest ones, to be competitive in one dimension may inevitably lag in others. For example, MME-Realworld \cite{data:MME-RealWorld}, as the most difficult benchmark, shows a Fallacy close to 0.5; and as the least redundant, it has relatively lagged Diversity because samples without Redundancy share certain commonalities. \CLA{There is no best benchmark, only the most suitable one.} A lower score does not signify a benchmark is inferior\footnote{This is why we refrain from ranking the overall Information Density.}. MLLM developers can combine their specific needs to select the dimensions they want to verify, thus testing the most suitable benchmarks.

% \begin{figure}[tbph]
% \centering
% \includegraphics[width = \linewidth]{vis/Main-Time.pdf}
% \caption{Time.}
% \vspace{-2mm}
% \label{fig:time}
% \end{figure}

\begin{figure}[tb]
\centering
\begin{minipage}[]{\linewidth}
  \centering
  \centerline{\includegraphics[width = \linewidth]{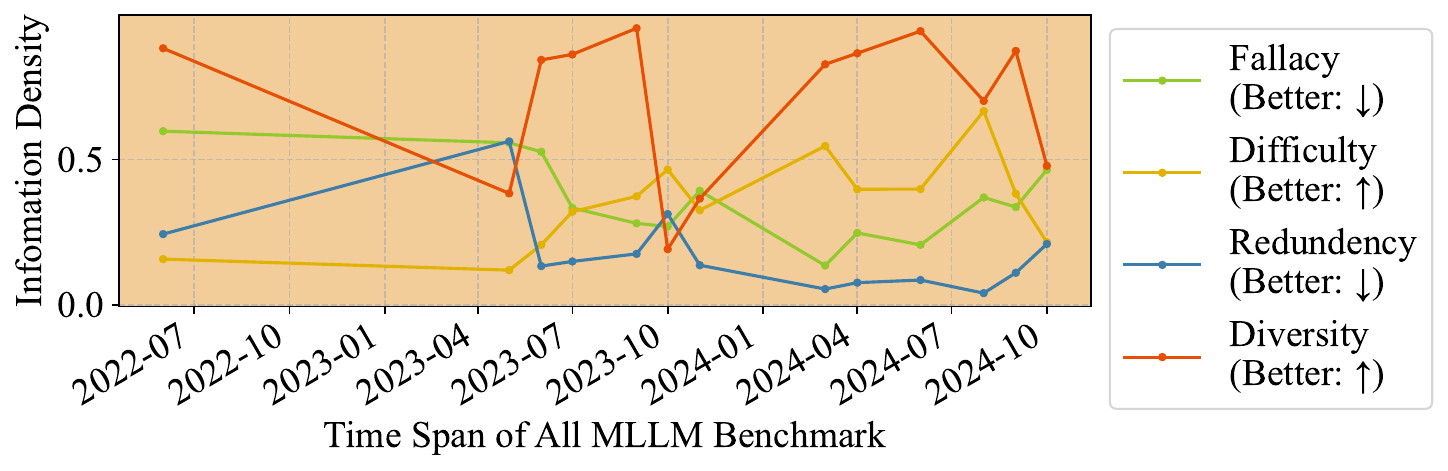}}
%  \centerline{(a)}\medskip
\end{minipage}
\begin{minipage}[]{\linewidth}
  \centering
  \centerline{\includegraphics[width = \linewidth]{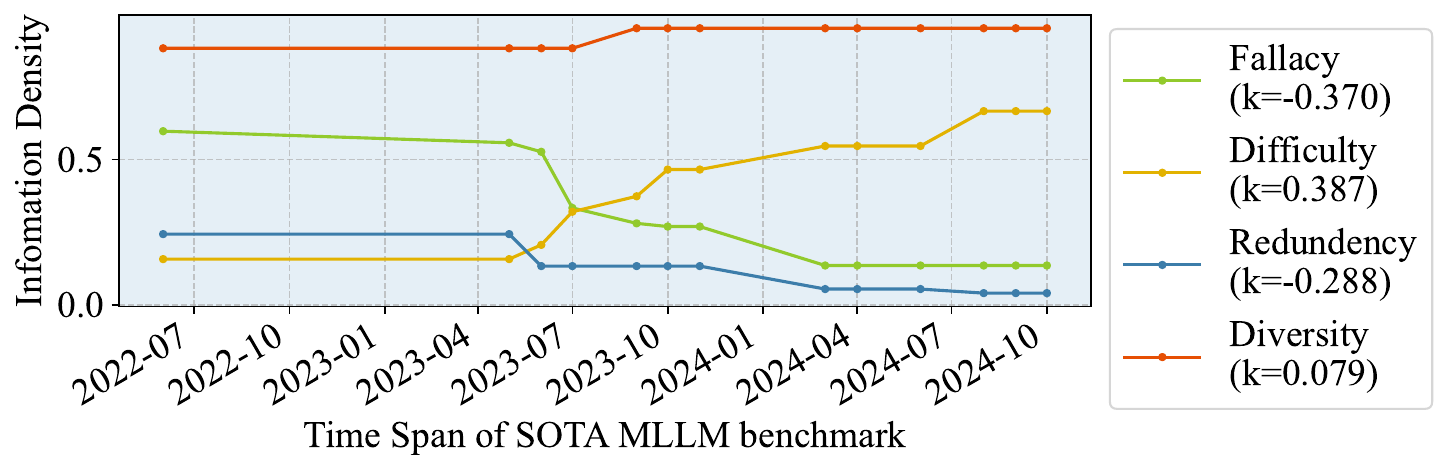}}
%  \centerline{(a)}\medskip
\end{minipage}
\vspace{-3mm}
\caption{Benchmark development trend over time. Fallacy and Difficulty improved the fastest, followed by Redundancy, and Diversity remained almost unchanged. [Keys: \CLA{All}; \CLB{SOTA}]}
\vspace{-6mm}
\label{fig:time}
\end{figure}

\subsection{Inter-Dimensional and Time Analysis}

Beyond only characterizing `How much insight a benchmark provides’, but also providing insights for the benchmark itself, we conducted an in-depth analysis. Figure \ref{fig:corrmat} shows the correlation between the various sub-dimensions of information density, also represented by the average of SRCC and PLCC.
Generally, the inter-dimensional correlation is low, demonstrating the rationality of the dimensional division. Meanwhile, only Difficulty exhibits a strong inner-dimensional correlation, while others are weakly correlated below 0.32.
This low correlation indicates that although MLLM developers can focus on specific dimensions, benchmark developers need to test across all dimensions. 
Figure \ref{fig:time} illustrates the trend of benchmark information density and calculates the slope $k$ for each indicator over time. The decrease in Fallacy and the increase in Difficulty demonstrate progress in these two dimensions; Redundancy follows, as previous developers paid insufficient attention to it until the MMstar highlighted this issue, but it has gradually been addressed; Unfortunately, there has been no significant change in Diversity. Therefore, we offer the following three suggestions for future benchmarks:

\begin{itemize}
    \item \CLA{Balancing Fallacy and Difficulty}: The optimal questions in the benchmark should be common sense for humans, but the MLLM cannot answer correctly. Compared with solving questions that human experts don't know, it can better expand the knowledge boundary of MLLM.
    \item \CLA{Performing verification of Redundancy}: Before releasing the benchmark, make a secondary verification without certain information and delete the correct samples. This little overhead can greatly improve its usability.
    \item \CLA{Improving Diversity (most important)}: With the rapid development of MLLM, its progress is far less than the other three. Benchmark developers need to optimize data collection and avoid similar images and templated text.
\end{itemize}

\section{Conclusion}

This paper introduces the Information Density principle for evaluating multimodal benchmarks, emphasizing its importance for MLLM development through an information entropy theoretical framework with four independent dimensions—Fallacy, Difficulty, Redundancy, and Diversity. A comprehensive evaluation of 19 mainstream benchmarks proved their defects due to these issues. For MLLM developers, this finding provides insights for selecting the most suitable benchmark. This paper also presents a Human-Model-Data evaluation pipeline as a validation tool for benchmark developers, thus contributing to the advancement of MLLM evaluation practices.

\clearpage
% \maketitlesupplementary
\appendix

\section{Limitation and Broader Impact}

\textbf{[Limitation]}: As the title suggests, information density is a principle rather than a complete suite. The main contribution of this paper lies in proposing the concept of information density and establishing a theoretical foundation, thereby demonstrating the deficiencies of the current benchmarks across multiple dimensions. However, as an evaluation toolkit, this pipeline is merely an initial attempt for further optimization. \textbf{This primarily reflects the trade-off between performance and cost, as the Human-Model-Data architecture is challenging to achieve both simultaneously.} For benchmark developers, organizing human annotations or having GPU resources is required to accurately assess information density; the credibility of directly evaluating data still needs improvement. Considering the low correlation between the four dimensions and the alignment of benchmark performance with human cognition, the information density is proved reasonable. Therefore, a more reliable evaluation mechanism for benchmarks can be developed based on this principle in the future.
\\
\textbf{[Broader Impact 1]}: For MLLM developers, this paper provides a leaderboard that serves as a basis for selecting benchmarks. Since hundreds of MLLM benchmarks coexist, developers are confused in determining which benchmarks to test. With the help of the information density leaderboard, they can choose dimensions based on their specific needs, such as testing the upper limits of MLLM capabilities (Difficulty) or ensuring reliable assessments (Fallacy), thus conducting an effective validation. This significantly constrains the costs of MLLM evaluations by improving its accuracy and efficiency testing, providing developers with more insights, and advancing the progress of the entire research community.
\\
\textbf{[Broader Impact 2]}: For benchmark developers, this paper offers a metric to ensure the effectiveness of their data. Over the past two years, benchmarks as an evaluation mechanism, have not been evaluated. A better benchmark is regarded as a larger data scale, higher image resolution, or longer texts. With the help of the information density principle, benchmark developers can first verify their benchmarks, such as determining if the questions are sufficiently challenging (Difficulty) and whether the samples cover various fields (Diversity). After ensuring the superiority of all dimensions (or the specific dimensions of interest to developers), they can formally release the benchmarks. This indicates directions for the optimization of future benchmarks, thus driving the evolvement of MLLMs.

\begin{figure}[t]
\centering
\begin{minipage}[]{\linewidth}
  \centering
  \centerline{\includegraphics[width = \textwidth]{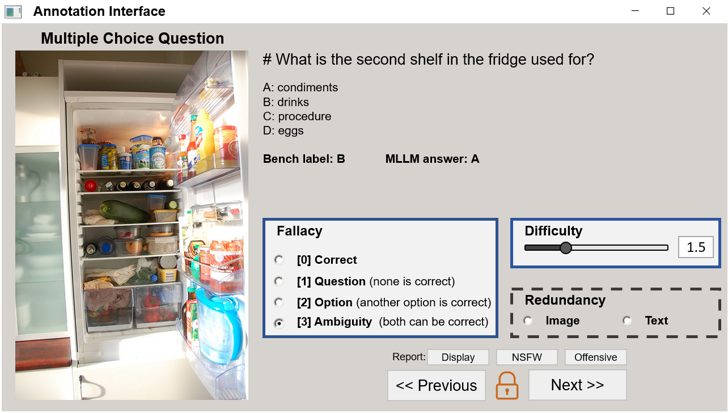}}\medskip
\end{minipage}
\begin{minipage}[]{\linewidth}
  \centering
  \centerline{\includegraphics[width = \textwidth]{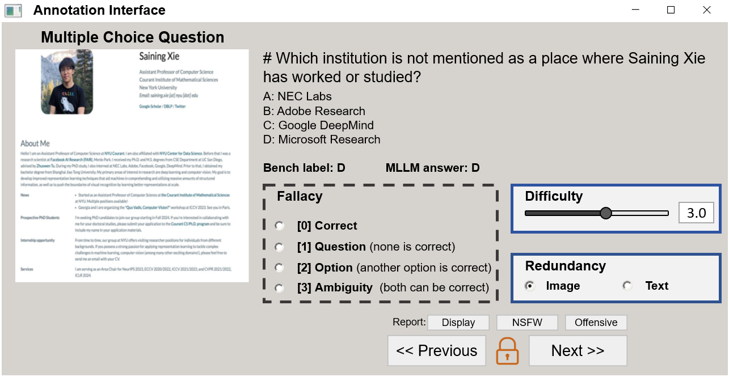}}\medskip
\end{minipage}
\vspace{-2mm}

\caption{User interface in labeling two different samples. Correctly answered samples are mandatory on Fallacy and Difficulty annotations; while incorrect samples need Difficulty and Redundancy. Users can click Unlock to operate non-mandatory annotations. Diversity will be labeled after viewing all samples.}
%\vspace{-2mm}
\label{fig:supp-interface}
\end{figure}

\begin{table}[tb]
\centering
    % \caption{Absolute robustness comparison between \textit{GPT-4o} and \textit{human} (\textbf{left/right}). Evaluated by 3 tasks, 3 strength, 7 steps, and 7 groups. As the R-bench champion, \textit{GPT-4o} still lags behind \textit{human} across the board.  \CLB{Orange}/\CLA{Blue} denote \textit{GPT-4o} performance below 90\% or above 98\% of \textit{humans}.}
    \caption{Difficulty result from the Data Eval. Structure (for image), Grammar (for text), Option (for text), and Region (for image+text). Their combination can represent the difficulty.}
    \label{tab-supp:difficulty}
    \vspace{-8pt}
    \renewcommand\arraystretch{1.4}
    \renewcommand\tabcolsep{7.5pt}
    \belowrulesep=0pt\aboverulesep=0pt
    \resizebox{\linewidth}{!}{
    \begin{tabular}{l|cccc}
    \toprule
    Benchmark               & Structure & Grammar & Option & Region \\ \midrule
    A-okvqa        & 0.062     & 0.421   & 0.390  & 0.071  \\
    POPE           & 0.056     & 0.327   & 0.719  & 0.071  \\
    MME            & 0.035     & 0.332   & 0.719  & 0.067  \\
    MMBench-v1.0   & 0.036     & 0.434   & 0.446  & 0.066  \\
    SEEDBench      & 0.033     & 0.451   & 0.453  & 0.068  \\
    Q-Bench        & 0.026     & 0.481   & 0.526  & 0.076  \\
    HallusionBench & 0.039     & 0.339   & 0.719  & 0.074  \\
    SEEDBench2     & 0.029     & 0.452   & 0.455  & 0.077  \\
    MMStar         & 0.053     & 0.356   & 0.544  & 0.067  \\
    MMBench-v1.1   & 0.039     & 0.438   & 0.454  & 0.066  \\
    RealWorldQA    & 0.021     & 0.432   & 0.696  & 0.079  \\
    SEEDBench2+    & 0.059     & 0.378   & 0.492  & 0.076  \\
    MMMB           & 0.037     & 0.445   & 0.555  & 0.068  \\
    A-Bench        & 0.029     & 0.433   & 0.484  & 0.074  \\
    TaskMeAnything & 0.018     & 0.485   & 0.444  & 0.075  \\
    MME-Realworld  & 0.031     & 0.340   & 0.413  & 0.085  \\
    HR-Bench       & 0.017     & 0.481   & 0.539  & 0.093  \\
    R-Bench        & 0.057     & 0.450   & 0.624  & 0.075  \\
    NaturalBench   & 0.051     & 0.457   & 0.580  & 0.076 \\ \bottomrule
    \end{tabular}
    
}
\end{table}

\begin{table}[tb]
\centering
    \caption{The intersection of Junior and Ambiguous sub-dimension in Difficulty. Their negligible intersection ratio proves the rationality of adding them directly to our pipeline.}
    \label{tab-supp:nointer}
    \vspace{-8pt}
    \renewcommand\arraystretch{1.4}
    \renewcommand\tabcolsep{6.5pt}
    \belowrulesep=0pt\aboverulesep=0pt
    \resizebox{\linewidth}{!}{
\begin{tabular}{c|c|c|c}
\toprule
A-okvqa   & POPE         & MME            & MMBench-v1.0  \\  \hdashline
0.004     & 0.000        & 0.000          & 0.002         \\ \hline
SEEDBench & Q-Bench      & HallusionBench & SEEDBench2    \\ \hdashline
0.012     & 0.014        & 0.002          & 0.013         \\ \hline
MMStar    & MMBench-v1.1 & RealWorldQA    & SEEDBench2+   \\ \hdashline
0.025     & 0.007        & 0.005          & 0.024         \\ \hline
MMMB      & A-Bench      & TaskMeAnything & MME-Realworld \\ \hdashline
0.002     & 0.018        & 0.014          & 0.029         \\ \hline
HR-Bench  & R-Bench      & NaturalBench   &               \\ \hdashline
0.009     & 0.010        & 0.001          &              \\ \bottomrule
\end{tabular}}
\end{table}

\section{Human Expert Labeling}

After MLLM conducted inference on the benchmark, we organized five Ph. D. candidates (with backgrounds in computer science, sociology, psychology, and AI4Science) as a panel to verify whether the original annotations of these samples were correct, as shown in Figure \ref{fig:supp-interface}. The broad knowledge background of these subjects ensures that they can understand the specialized content in the vast majority of samples, ensuring high-quality annotations. Each subject scores the dataset independently, following a randomly shuffled sequence, to ensure the continuity of annotation standards and to prevent fatigue. First, each question must be annotated for Difficulty. Secondly, for questions answered incorrectly, it needs to be ensured that the question is indeed challenging, rather than having an error in the original annotation. Finally, for questions answered correctly, it needs to be verified whether the answer could still be correct if the image/text were missing, i.e., Redundancy. After completing the annotations, to collect total scores, the rules of Fallacy are classified into three subcategories as described in the text; the average difficulty and redundancy of all questions will be used as labels for the entire dataset; Diversity is evaluated after viewing the entire dataset.

\section{Detail about Four Dimensions}

This section mainly explains the theoretical derivation in the main text in detail and presents additional results.

\begin{table}[t]
\centering
    % \caption{Absolute robustness comparison between \textit{GPT-4o} and \textit{human} (\textbf{left/right}). Evaluated by 3 tasks, 3 strength, 7 steps, and 7 groups. As the R-bench champion, \textit{GPT-4o} still lags behind \textit{human} across the board.  \CLB{Orange}/\CLA{Blue} denote \textit{GPT-4o} performance below 90\% or above 98\% of \textit{humans}.}
    \caption{Merging the opinion of five subjects into the final label in the fallacy dimension. The definition of label 0,1,2,3 follows Figure \ref{fig:supp-interface}. X stands for any number, repeatable; and A/B/C denotes different numbers in 1,2,3.}
    \label{tab:supp-fallacy}
    \vspace{-8pt}
    \renewcommand\arraystretch{1.4}
    \belowrulesep=0pt\aboverulesep=0pt
    \resizebox{\linewidth}{!}{
\begin{tabular}{l|c|p{6cm}}
\toprule
Answer & Label & \multicolumn{1}{c}{Reason}                                                                                                                       \\ \midrule
000XX  & 0     & The knowledge of the two subjects is not broad enough.                                                                            \\ \hdashline
0011X  & 0     & Most subjects cannot understand the question, which means it is a proprietary field. Trust the knowledge of the two experts. \\ \hdashline
00AAB  & A     & The minority obeys the majority.                                                                                                  \\ \hdashline
00123  & 2     & Some choose all while some discard the original. All options are correct but the new option has high confidence.                  \\ \hdashline
0AABC  & A     & The minority obeys the majority.                                                                                                  \\ \hdashline
AAAXX  & A     & The minority obeys the majority.                                                                                                  \\ \hdashline
X1122  & 2     & An option called `I can't tell', which just means (1), but has a higher priority.                                   \\ \hdashline
X1133  & 3     & Options are based on the same premise and are either all true or all false.                                                      \\ \hdashline
X2233  & 2     & New option has higher confidence.            \\ \bottomrule
\end{tabular}}
\vspace{-3mm}
\end{table}

\subsection{Difficulty}

In Data Eval, the low-level features extracted by each benchmark are shown in Table \ref{tab-supp:difficulty}. Overall, the correlation between the first three dimensions and Model Eval is about 0.3, while a correlation of 0.5 can be achieved using only the Region dimension. Therefore, these features are sufficient to fit the benchmark difficulty obtained by Model Eval.

In addition, in Model Eval, to avoid analyzing each sample one by one, we directly add the proportion of samples with incorrect MLLM answers and the proportion of samples with correct but ambiguous answers. This mechanism is proven to be reasonable in Table\ref{tab-supp:nointer}, because the intersection ratio of these two categories is less than 0.03, which is almost non-existent.

\subsection{Fallacy}

When merging the annotations of the five subjects, to ensure rigor, we did not adopt the majority rule simply but developed a standard process based on the characteristics of each Fallacy. First, if more than three people think the original annotation is correct, or two people think it is correct and the other two think all the options are wrong, it means that this tends to be a professional issue, two experts understand it and the other two do not have knowledge in this field, so they still think the annotation is correct. In other cases, the majority rule is basically followed, and when there is a tie, the priority order is $2>3>1$. The specific reasons are shown in Table \ref{tab:supp-fallacy}.

\begin{table}[tb]
    \caption{Redundancy-related attributes of each benchmark. \faTimesCircle denotes this benchmark is not applicable to text redundancy.}
    \label{tab-supp:redundancy}
    \vspace{-8pt}
    \renewcommand\arraystretch{1.4}
    \renewcommand\tabcolsep{6.6pt}
    \belowrulesep=0pt\aboverulesep=0pt
    \resizebox{\linewidth}{!}{
\begin{tabular}{l|cccc}
\toprule
Benchmark               & Img token & Txt token    & Options   & Txt \\ \midrule
A-okvqa        & 167      & 23.24    & 4     &     \\
POPE           & 167      & 21.53    & 2     & \faTimesCircle   \\
MME            & 167      & 24.77    & 2     & \faTimesCircle   \\
MMBench        & 167      & 40.50    & 3.781 &     \\
SEEDBench      & 167      & 32.33    & 3.984 &     \\
Q-Bench        & 167      & 19.45    & 2.844 &     \\
HallusionBench & 167      & 26.57 & 2     & \faTimesCircle   \\
SEEDBench2     & 167      & 31.92    & 3.985 &     \\
MMStar         & 167      & 40.25    & 3.823 &     \\
MMBench-v1.1   & 167      & 39.30     & 3.796 &     \\
RealWorldQA    & 167      & 27.96 & 2.750  &     \\
SEEDBench2+    & 167      & 39.80    & 4     &     \\
MMMB           & 167      & 24.86    & 3.097 &     \\
A-Bench        & 167      & 28.47    & 3.166 &     \\
TaskMeAnything & 167      & 28.91    & 4     &     \\
MME-Realworld  & 167      & 55.92    & 4.999 &     \\
HR-Bench       & 167      & 34.80    & 4     &     \\
R-Bench        & 167      & 28.21 & 3.036 &     \\
NaturalBench   & 167      & 27.93    & 2     & \faTimesCircle  \\ \bottomrule
\end{tabular}}
\vspace{-3mm}
\end{table}

\begin{table}[]
    \caption{Diversity result form the Data Eval. a1-a5 denote Luminance, Contrast, Chrominance, Blur, and Spatial Information.}
    \label{tab-supp:diversity}
    \vspace{-8pt}
    \renewcommand\arraystretch{1.4}
    \renewcommand\tabcolsep{6.5pt}
    \belowrulesep=0pt\aboverulesep=0pt
    \resizebox{\linewidth}{!}{
\begin{tabular}{l|ccccc}
\toprule
Benchmark               & a1 & a2 & a3 & a4 & a5 \\ \midrule
A-okvqa        & 0.960      & 0.919      & 0.941      & 0.847       & 0.969       \\
POPE           & 0.973      & 0.911      & 0.946      & 0.829      & 0.968      \\
MME            & 0.964      & 0.952      & 0.938      & 0.844      & 0.974      \\
MMBench        & 0.947       & 0.898      & 0.925      & 0.798      & 0.979      \\
SEEDBench      & 0.966      & 0.959      & 0.914      & 0.855      & 0.960      \\
Q-Bench        & 0.961      & 0.892      & 0.901      & 0.689       & 0.978      \\
HallusionBench & 0.835       & 0.877      & 0.845      & 0.789      & 0.918      \\
SEEDBench2     & 0.939      & 0.861      & 0.957      & 0.782       & 0.954       \\
MMStar         & 0.856      & 0.896      & 0.903       & 0.786      & 0.974      \\
MMBench-v1.1   & 0.954      & 0.913      & 0.912      & 0.836      & 0.976      \\
RealWorldQA    & 0.855      & 0.945      & 0.823      & 0.688      & 0.881      \\
SEEDBench2+    & 0.798      & 0.931      & 0.902      & 0.811      & 0.945      \\
MMMB           & 0.926      & 0.899      & 0.933      & 0.835      & 0.966      \\
A-Bench        & 0.938       & 0.926      & 0.909      & 0.784      & 0.962      \\
TaskMeAnything & 0.909      & 0.956      & 0.924      & 0.686      & 0.926       \\
MME-Realworld  & 0.806      & 0.840      & 0.871      & 0.732      & 0.962      \\
HR-Bench       & 0.864      & 0.955      & 0.927      & 0.806      & 0.944      \\
R-Bench        & 0.906      & 0.955      & 0.890      & 0.684      & 0.930      \\
NaturalBench   & 0.936      & 0.929      & 0.917      & 0.832      & 0.960 \\ \bottomrule    
\end{tabular}}
\vspace{-3mm}
\end{table}

\subsection{Redundancy}

Redundancy requires proportional integration of image and text components on one hand and clarity on when it is meaningful to calculate redundancy on the other. Both aspects are explained in Table \ref{tab-supp:redundancy}.
Firstly, we use the number of tokens required by the MLLM when processing images and text to represent their respective code weights. For images, the model typically downsamples them to 224x224 for input, which corresponds to approximately 167 tokens in the architectures of the LLaVA and Qwen-VL series. Given that there are rarely smaller images in mainstream benchmarks, we directly set this as the token count for images.
For the text, we sequentially calculate the token count for each text input in the dataset and then take the average (excluding prefixes, suffixes, hints, etc.). This attribute primarily depends on the average word count of the benchmark. Finally, we calculate the effective number of options available in the benchmark. When there are only two options, calculating the text redundancy is meaningless because if the question is unknown, guessing between the two options is not a valid approach (in contrast, having more options allows for the use of the elimination method), so redundancy is not calculated. However, when there are enough options, for example, at least 2.75 as shown in Table \ref{tab-supp:redundancy}, using five random seeds for testing indicates that the samples that answered all correctly had almost no chance of guessing correctly. Meanwhile, image redundancy does not have such constraints and can be calculated in any scenario.

\subsection{Diversity}

First, we show the five features of the image in Data Eval mode, as shown in Table \ref{tab-supp:diversity}. As the feature distribution in the main text, their linear combination can be used to fit Model Eval. Next, we derive the data filtering mechanism in the main text, namely the $SIM(\cdot)$ function, in detail:
\textbf{Text/Image Embedding Computation.} 
Given a dataset of textual or visual data \( \mathcal{D} = \{x_1, x_2, \dots, x_N\} \), each sample \( x_i \) is encoded using a pre-trained embedding model \( f_{\theta} \), (e.g., (vision) transformer), to obtain the embeddings:

\begin{equation}
    \mathbf{e}_i = f_{\theta}(x_i), \quad \forall i \in \{1, 2, \dots, N\}
\end{equation}
where $\mathbf{e}_i \in \mathbb{R}^d$ represents the $d$-dimensional vector embedding of text $x_i$. The complete embedding matrix is:
\begin{equation}
    \mathbf{E} = [\mathbf{e}_1, \mathbf{e}_2, \dots, \mathbf{e}_N]^\top \in \mathbb{R}^{N \times d}
\end{equation}
\textbf{K-means Clustering.} 
To group similar texts/images, we apply K-means clustering on the embeddings $\mathbf{E}$ by minimizing the intra-cluster Euclidean distance:

\begin{equation}
    \min_{\mathcal{C}, \mathbf{\mu}} \sum_{k=1}^{K} \sum_{\mathbf{e}_i \in \mathcal{C}_k} \|\mathbf{e}_i - \mathbf{\mu}_k\|^2
\end{equation}
where:
\begin{itemize}
    \item $\mathcal{C}_k$ represents the $k$-th cluster, ensuring $\bigcup_{k=1}^{K} \mathcal{C}_k = \mathcal{D}$.
    \item $\mathbf{\mu}_k = \frac{1}{|\mathcal{C}_k|} \sum_{\mathbf{e}_i \in \mathcal{C}_k} \mathbf{e}_i$ is the centroid of the $k$-th cluster.
\end{itemize}
Each text sample $x_i$ is assigned to a cluster based on the minimum distance criterion:

\begin{equation}
    c_i = \arg\min_k \|\mathbf{e}_i - \mathbf{\mu}_k\|^2, \quad k \in \{1, 2, \dots, K\}
\end{equation}
\textbf{Intra-cluster Sorting.} 
Within each cluster $\mathcal{C}_k$, we compute the pairwise \textbf{cosine similarity}:

\begin{equation}
    S_{ij} = \frac{\mathbf{e}_i \cdot \mathbf{e}_j}{\|\mathbf{e}_i\| \|\mathbf{e}_j\|}, \quad \forall \mathbf{e}_i, \mathbf{e}_j \in \mathcal{C}_k
\end{equation}
We then compute the \textbf{average similarity} for each sample:

\begin{equation}
    \bar{S}_i = \frac{1}{|\mathcal{C}_k|} \sum_{\mathbf{e}_j \in \mathcal{C}_k} S_{ij}, \quad \forall \mathbf{e}_i \in \mathcal{C}_k
\end{equation}
Finally, the samples are sorted in descending order from $\bar{S}_i$:

\begin{equation}
    \mathcal{C}_k^{\text{sorted}} = \text{argsort}(-\bar{S}_i)
\end{equation}
\textbf{Semantic Deduplication.} 
We perform \textbf{semantic deduplication} by computing cosine similarity between all pairs:

\begin{equation}
    S_{ij} = \frac{\mathbf{e}_i \cdot \mathbf{e}_j}{\|\mathbf{e}_i\| \|\mathbf{e}_j\|}
\end{equation}
We define a threshold $\tau$, and if: $S_{ij} > \tau$, \text{then texts } $x_i$ \text{ and } $x_j$ \text{ are considered semantically similar.}
Using a \textbf{greedy deduplication strategy}: Traverse the dataset and remove any $x_j$ where $S_{ij} > \tau$ for any existing $x_i$. The final unique dataset is denoted as:

\begin{equation}
    {\rm SIM} (\mathcal{D}) = \{x_i \mid \forall x_j, S_{ij} \leq \tau\}
\end{equation}

\begin{figure*}
\centering
\begin{minipage}[]{0.24\linewidth}
  \centering
  \centerline{\includegraphics[width = \textwidth]{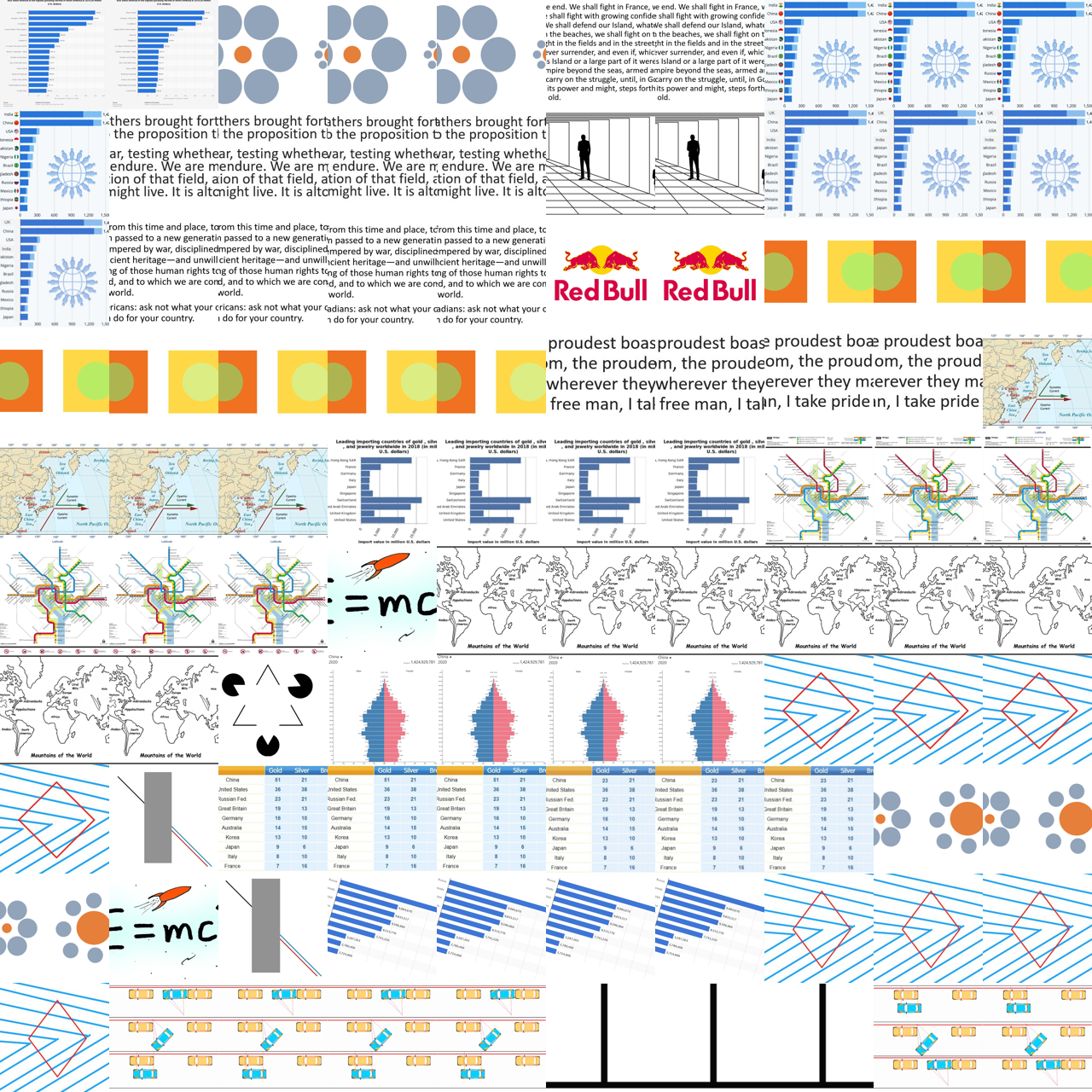}}
  \centerline{HallusionBench (0.191)}\medskip
\end{minipage}
\begin{minipage}[]{0.24\linewidth}
  \centering
  \centerline{\includegraphics[width = \textwidth]{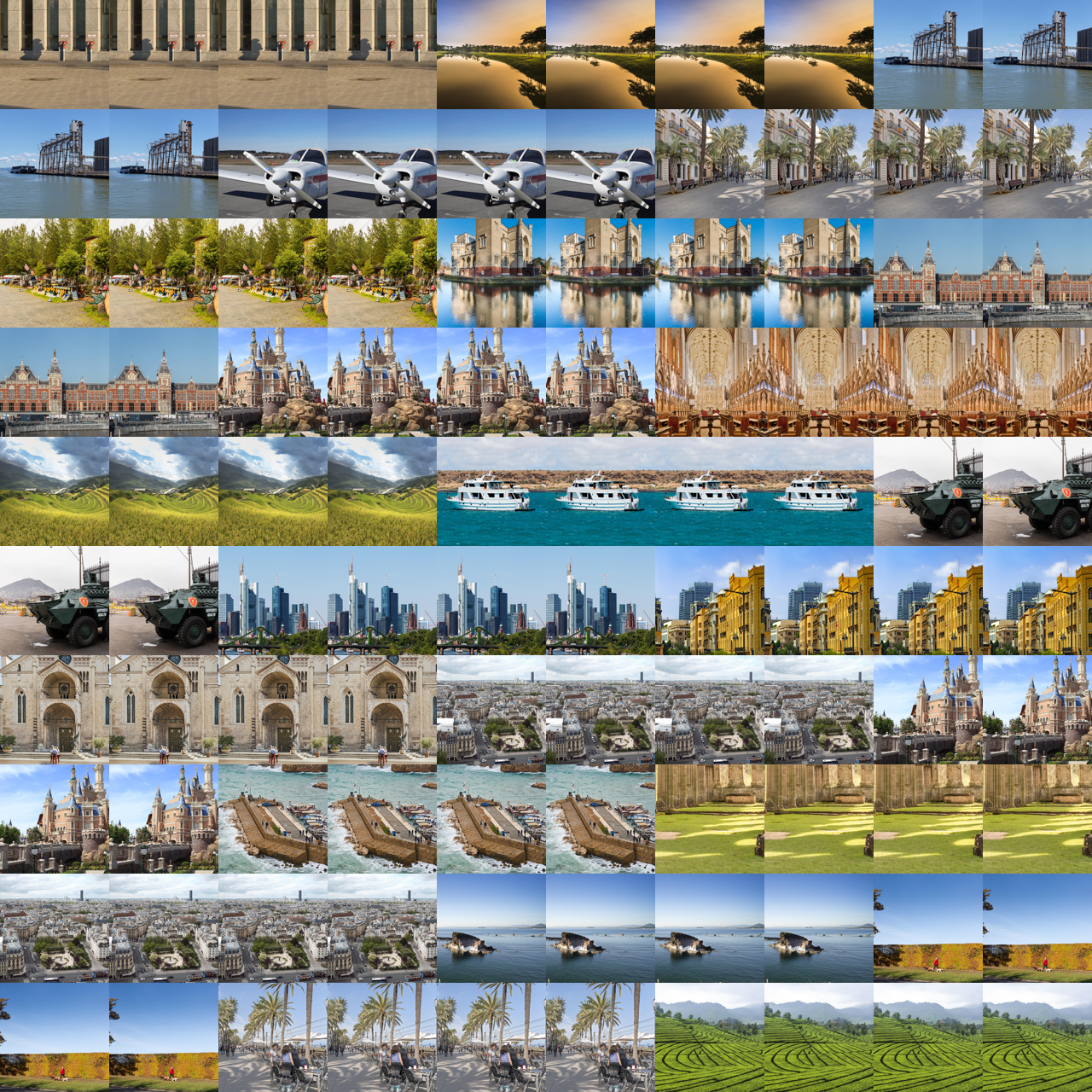}}
  \centerline{HR-Bench (0.205)}\medskip
\end{minipage}
\begin{minipage}[]{0.24\linewidth}
  \centering
  \centerline{\includegraphics[width = \textwidth]{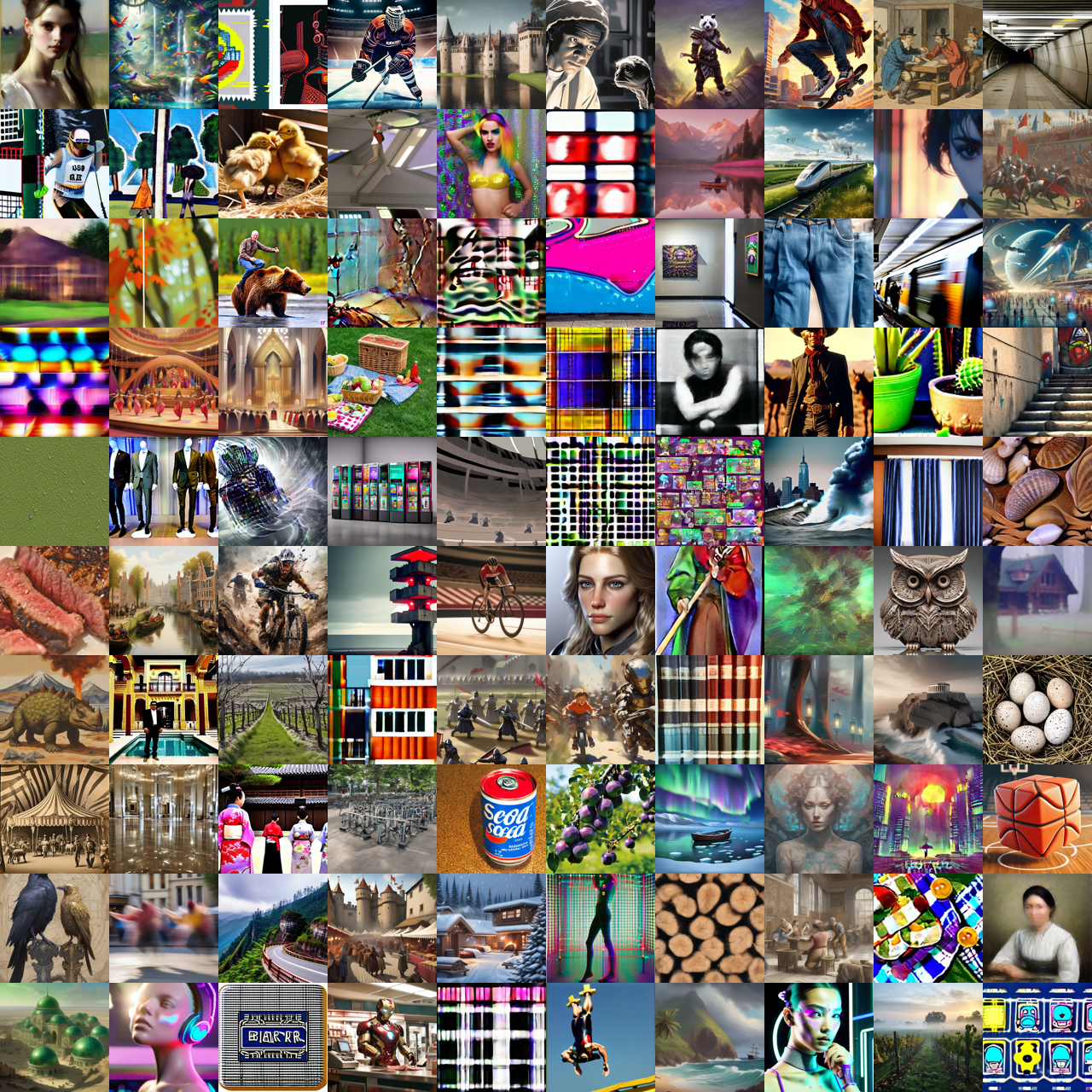}}
  \centerline{A-Bench (0.941)}\medskip
\end{minipage}
\begin{minipage}[]{0.24\linewidth}
  \centering
  \centerline{\includegraphics[width = \textwidth]{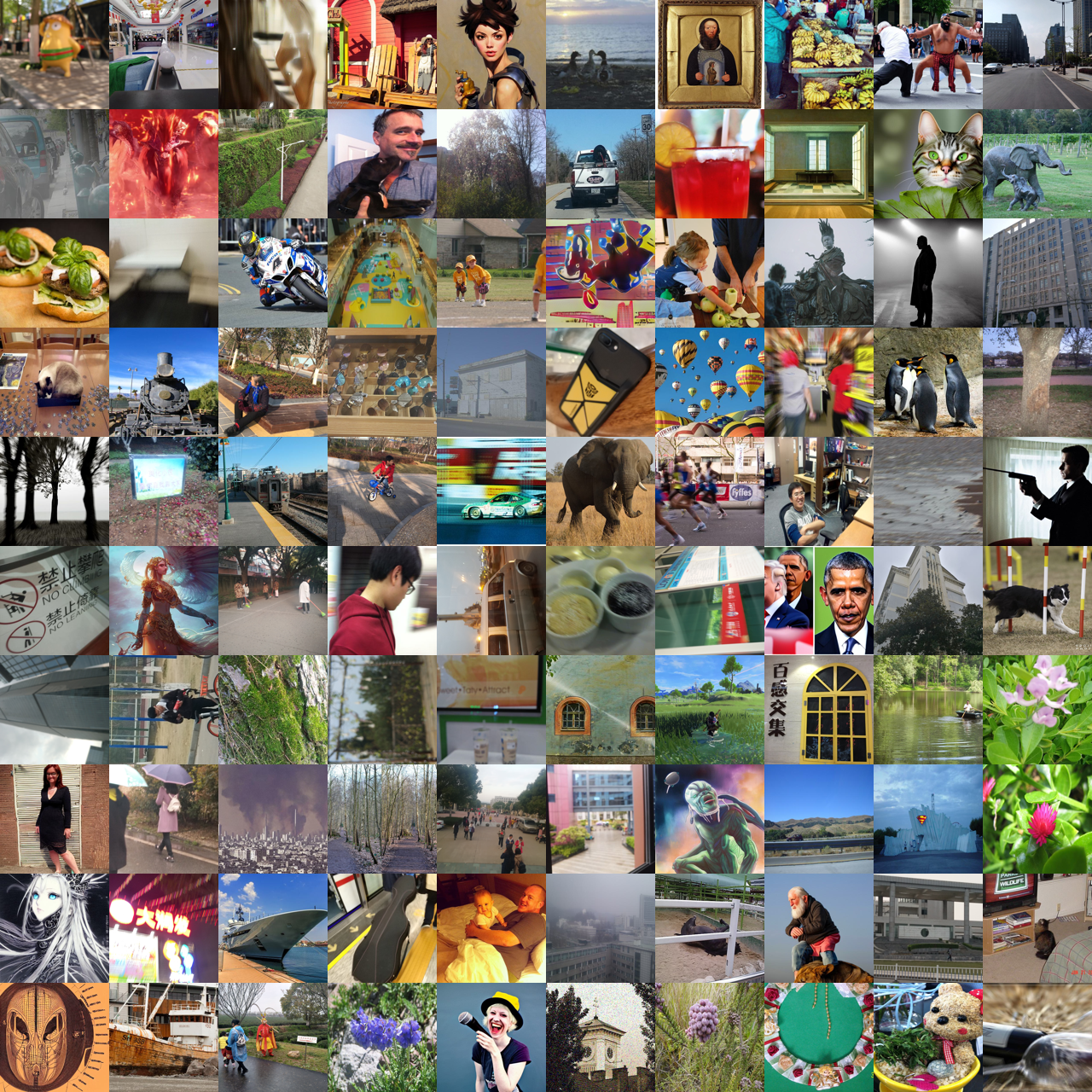}}
  \centerline{Q-Bench (0.951)}\medskip
\end{minipage}

\vspace{-2mm}

\caption{Example of benchmarks with high/low image diversity, represented by 10$\times$10 samples with image diversity score.}
%\vspace{-2mm}
\label{fig:supp-div-img}
\end{figure*}

\begin{figure*}
\centering
\begin{minipage}[]{0.24\linewidth}
  \centering
  \centerline{\includegraphics[width = \textwidth]{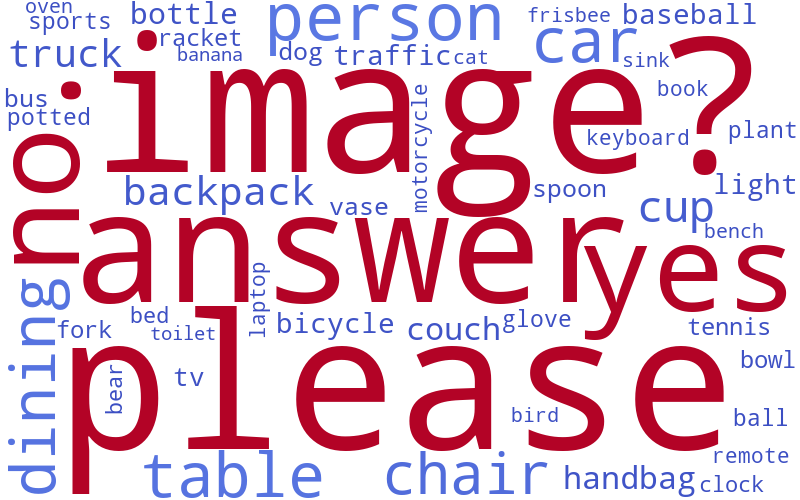}}
  \centerline{POPE (0.001)}\medskip
\end{minipage}
\begin{minipage}[]{0.24\linewidth}
  \centering
  \centerline{\includegraphics[width = \textwidth]{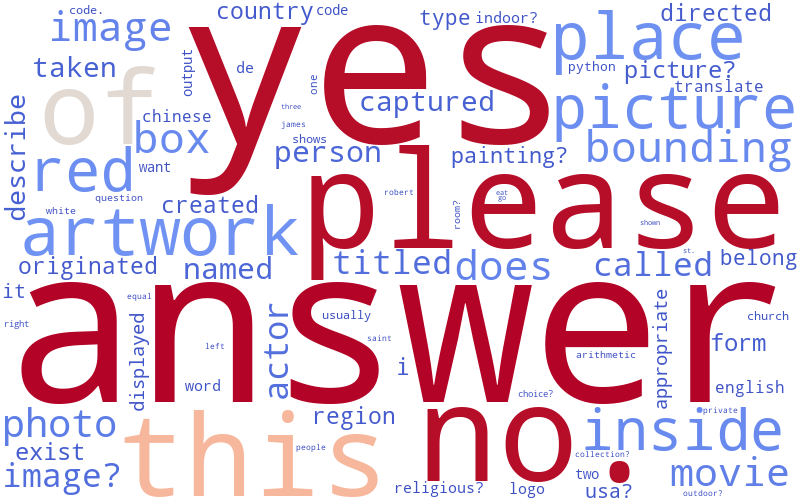}}
  \centerline{MME (0.113)}\medskip
\end{minipage}
\begin{minipage}[]{0.24\linewidth}
  \centering
  \centerline{\includegraphics[width = \textwidth]{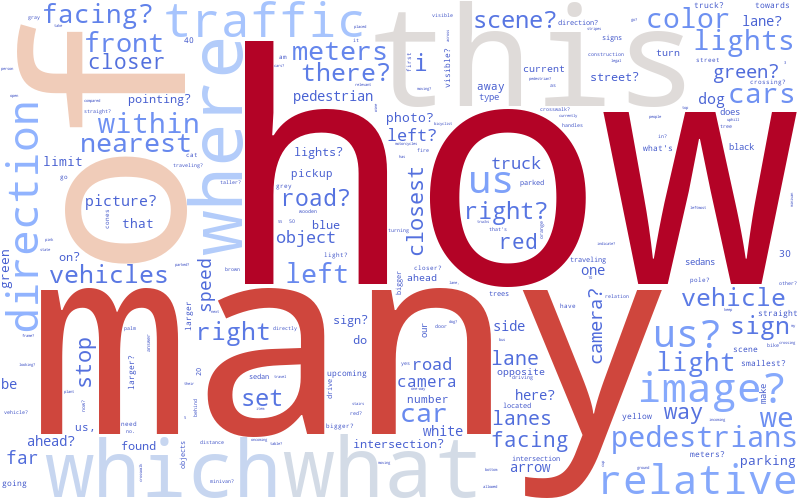}}
  \centerline{RealWorldQA (0.796)}\medskip
\end{minipage}
\begin{minipage}[]{0.24\linewidth}
  \centering
  \centerline{\includegraphics[width = \textwidth]{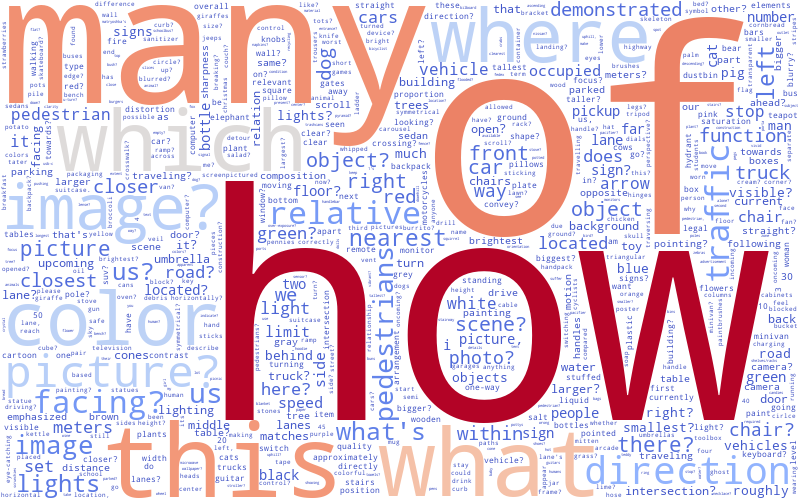}}
  \centerline{R-Bench (0.799)}\medskip
\end{minipage}

\vspace{-2mm}

\caption{Example of benchmarks with high/low text diversity, represented by word-clouds with text diversity score.}
%\vspace{-2mm}
\label{fig:supp-div-txt}
\end{figure*}

\section{Example of High\&Low Diversity Dataset}

Due to the page limitations, only attributes related to individual samples were presented in the main text. Here, we supplement the visualization of Diversity, which is determined by multiple samples collectively. Figure \ref{fig:supp-div-img} illustrates examples of image diversity, featuring two uniform examples and two richer examples. It is evident that HallusionBench and HR-Bench often repeat inquiries about single images, and the styles are relatively fixed as screen content and landscape; in contrast, A-Bench and Q-Bench do not have repeated images and cover various content, including portraits, objects, landscapes, and AI-generated images.

Figure \ref{fig:supp-div-txt} showcases textual diversity with both uniform and rich examples. The questions in POPE and MME are quite templated, all phrased as `please answer XXX in yes or no', with a limited variety of vocabulary. Conversely, RealworldQA and R-Bench present various question formats starting with `How/What/Which', with a denser vocabulary distribution compared to the former. This demonstrates the consistency of the evaluation mechanism with human subjective cognition.

\section{Scope of Application}

This section clarifies the application scope of information density. For benchmarks that do not meet certain conditions, some dimensions may be impossible to calculate, or the results may not be fair and objective. (In other words, it is not that the benchmark design is poor, but rather that the specific task scenario it addresses necessitates such a setup.) Only benchmarks that satisfy the conditions can yield complete testing results.

In the main text, we propose three conditions that benchmarks suitable for information density must meet: multiple-choice, multi-modal, and multi-domain. The first two are easy to understand, while the specific definition of multi-domain is as follows:

Image: Benchmarks constrained at the feature level can be tested, such as those containing AI or user-generated images while incorporating different subjects, backgrounds, and styles. Conversely, benchmarks constrained at the task level cannot be tested, such as those involving tables or geometry questions, where each question has multiple sub-questions, making it impossible to evaluate difficulty and diversity.

Text: Benchmarks constrained at the domain level can be tested, such as those assessing low-level or high-level visual understanding, as this does not limit the format of the questions. Benchmarks constrained at the discipline level can also be tested, such as those in chemistry or medicine, which require fixed templates.

Here are some example benchmarks that help understand the applicability:
\begin{itemize}
    \item Non-testable: OCR-Bench (images consist entirely of documents), ScienceQA (text is entirely scientific).
    \item Testable: A-bench (images are all generated), Q-bench (text focuses on low-level aspects).
\end{itemize}

In summary, a benchmark is applicable if the content of the images is constrained, any text can be used as a question (you can ask any question about generated images), and vice versa. However, if one is fixed, the other will also be limited (geometry questions can only ask math-related aspects), making them non-testable.

\section{Disclaimer}

This work represents an initial study on the \textbf{information density of multimodal benchmarks}. 
We define four key dimensions -- Fallacy, Difficulty, Redundancy, and Diversity -- 
to characterize multimodal benchmarks and further introduce a set of factors within each dimension to enable quantitative assessment of these properties. 
Due to the scope of this work, the defined factors inevitably provide an incomplete and non-exhaustive characterization of each dimension, leaving room for future research to supplement.

Additionally, \textbf{this work does not involve praising or criticizing any multimodal benchmarks, nor is it a commercial leaderboard for ranking.}
Among the four dimensions, only the Fallacy score has an objective measure of quality (where lower values are preferable). 
The other three dimensions are used to represent the amount of information, instead of inherently superior or inferior values, whose appropriateness depends on the specific evaluation context. 
For instance, a benchmark designed to assess MLLMs in autonomous driving scenarios will naturally exhibit lower visual Diversity than a general-purpose multimodal benchmark. 
Similarly, evaluating MLLM proficiency across different capability levels requires selecting benchmarks with an appropriate degree of Difficulty.

Meanwhile, we \textbf{fully recognize the contributions} made by early benchmarks, despite their various shortcomings, and the importance of proprietary benchmarks, although their information is less than general ones. Our goal is to draw trends from early-stage benchmarks and current benchmarks and, thus, look forward to subsequent benchmarks in the future.

In conclusion, we advocate that future MLLM researchers and benchmark developers carefully consider their specific objectives and responsibly apply the \textbf{Information Density Principle} proposed in this work to guide benchmark selection and design.

{
    \small
    \bibliographystyle{ieeenat_fullname}
    \bibliography{main}
}

\end{document}